\PassOptionsToPackage{table}{xcolor}
\PassOptionsToPackage{pagebackref,breaklinks=true,colorlinks,citecolor=blue,urlcolor=blue,linkcolor=blue,bookmarks=false}{hyperref}
\PassOptionsToPackage{noabbrev,nameinlink,capitalize}{cleveref}

\documentclass[onecolumn, numbers]{april_aigc}

\usepackage{amsmath,amsfonts,amssymb,bm}
\usepackage{xspace}
\usepackage{float}
\usepackage{wrapfig}
\usepackage{tabularx}
\usepackage{array}
\usepackage{makecell}
\usepackage{colortbl}
\usepackage{enumitem}
\usepackage[normalem]{ulem}

\usepackage{amsmath,amsfonts,bm}

\def\eqref#1{equation~\ref{#1}}

\def\1{\bm{1}}

\DeclareMathAlphabet{\mathsfit}{\encodingdefault}{\sfdefault}{m}{sl}
\SetMathAlphabet{\mathsfit}{bold}{\encodingdefault}{\sfdefault}{bx}{n}

\definecolor{VLMNavy}{HTML}{081F5C}
\definecolor{VLMCobalt}{HTML}{334EAC}
\definecolor{VLMBlue}{HTML}{7096D1}
\definecolor{VLMSky}{HTML}{BAD6EB}
\definecolor{VLMFrost}{HTML}{D0E3FF}
\definecolor{VLMCream}{HTML}{F7F2EB}
\definecolor{VLMPaper}{HTML}{FFF9F0}
\definecolor{VLMViolet}{HTML}{5A4FCF}
\definecolor{VLMDeepBlue}{HTML}{1E3478}
\definecolor{VLANoResult}{RGB}{232,232,232}
\definecolor{green_code}{RGB}{55,126,34}

\newcommand{\methodlit}{\textcolor{VLMBlue}{\textbf{EASG}}\xspace}
\newcommand{\method}{EASG\xspace}
\newcommand{\tabitem}{\textbullet\hspace{0.35em}}
\newcommand{\pipeline}[1]{%
  \begin{tabular}[c]{@{}l@{}}#1\end{tabular}%
}
\newcolumntype{C}{>{\centering\arraybackslash}X}
\newcolumntype{Y}{>{\centering\arraybackslash}X}

\newtcolorbox{expquestion}[2]{
  enhanced,
  breakable,
  sharp corners,
  boxrule=0.45pt,
  leftrule=1.8pt,
  colframe=#1,
  colback=VLMPaper,
  colbacktitle=#1,
  coltitle=VLMNavy,
  title={#2},
  fonttitle=\bfseries\normalsize,
  fontupper=\normalsize,
  left=5pt,
  right=5pt,
  top=0pt,
  bottom=0pt,
  toptitle=2.4pt,
  bottomtitle=2.4pt,
  titlerule=0pt,
  before skip=\baselineskip,
  after skip=\baselineskip,
}

\newtcolorbox{expfinding}[1]{
  enhanced,
  sharp corners,
  colback=white,
  colframe=#1,
  coltext=VLMNavy,
  fontupper=\normalsize,
  width=\linewidth,
  boxrule=0pt,
  leftrule=1.4pt,
  rightrule=0pt,
  toprule=0pt,
  bottomrule=0pt,
  boxsep=0pt,
  left=6pt,
  right=3pt,
  top=1.0pt,
  bottom=0pt,
  before={\par\noindent},
  after={},
  before skip=\baselineskip,
  after skip=\baselineskip,
}

\usepackage[utf8]{inputenc}
\usepackage{url}
\usepackage{nicefrac}
\usepackage{fix-cm}
\usepackage{times}

\renewcommand{\textbf}[1]{{\bfseries #1}}
\usepackage{multicol}
\usepackage{multirow}
\usepackage{adjustbox}
\usepackage{pifont}
\usepackage{algorithm}
\usepackage{algorithmic}
\usepackage{listings}
\usepackage{titletoc}
\usepackage{etoolbox}
\usepackage{fancyhdr}

\makeatletter
\AfterEndEnvironment{algorithm}{\let\@algcomment\relax}
\AtEndEnvironment{algorithm}{\kern2pt\hrule\relax\vskip3pt\@algcomment}
\let\@algcomment\relax
\newcommand\algcomment[1]{\def\@algcomment{\footnotesize#1}}
\renewcommand\fs@ruled{\def\@fs@cfont{\bfseries}\let\@fs@capt\floatc@ruled
  \def\@fs@pre{\hrule height.8pt depth0pt \kern2pt}%
  \def\@fs@post{}%
  \def\@fs@mid{\kern2pt\hrule\kern2pt}%
  \let\@fs@iftopcapt\iftrue}
\makeatother

\definecolor{linkcolor}{named}{aprilblue}
\definecolor{urlcolor}{RGB}{255,105,180}
\definecolor{citecolor}{RGB}{66,168,235}
\definecolor{lightgray}{rgb}{0.8,0.8,0.8}
\definecolor{darkgreen}{rgb}{0.00,0.81,0.78}
\definecolor{gray_tab}{RGB}{220,220,220}
\definecolor{blue_tab}{RGB}{227,240,251}
\definecolor{oran_tab}{RGB}{252,242,237}
\definecolor{whit_tab}{RGB}{255,255,255}
\definecolor{codeblue}{RGB}{0,102,204}
\definecolor{codekw}{RGB}{128,0,128}

\hypersetup{colorlinks=true,linkcolor=linkcolor,urlcolor=urlcolor,citecolor=citecolor}
\AtEndPreamble{
  \usepackage[capitalize]{cleveref}
  \crefname{section}{Sec.}{Secs.}
  \Crefname{section}{Section}{Sections}
  \crefname{table}{Tab.}{Tabs.}
  \Crefname{table}{Table}{Tables}
  \crefname{equation}{Eq.}{Eqs.}
  \Crefname{equation}{Equation}{Equations}
  \crefname{figure}{Fig.}{Figs.}
  \Crefname{figure}{Figure}{Figures}
}

\titleformat*{\section}{\color{aprilblue}\Large\rmfamily\scshape\raggedright}
\titleformat*{\subsection}{\color{aprilblue}\large\rmfamily\scshape\raggedright}
\titleformat*{\subsubsection}{\color{aprilblue}\normalsize\rmfamily\scshape\raggedright}
\DeclareCaptionFormat{aprilcustom}{{\color{aprilblue}\sffamily\textbf{#1 #2}} #3}
\renewcommand{\authorformat}[2][]{{\rmfamily\bfseries #2$^{#1}$}}
\renewcommand{\title}[1]{%
  \def\titlelist{{\fontsize{20pt}{28pt}\selectfont
  \rmfamily\bfseries #1}}%
}

\newif\ifshowlogo
\showlogotrue
\newcommand{\insertlogo}{%
  \ifshowlogo
    \IfFileExists{assets/april_logo1.png}%
      {\includegraphics[height=0.68cm]{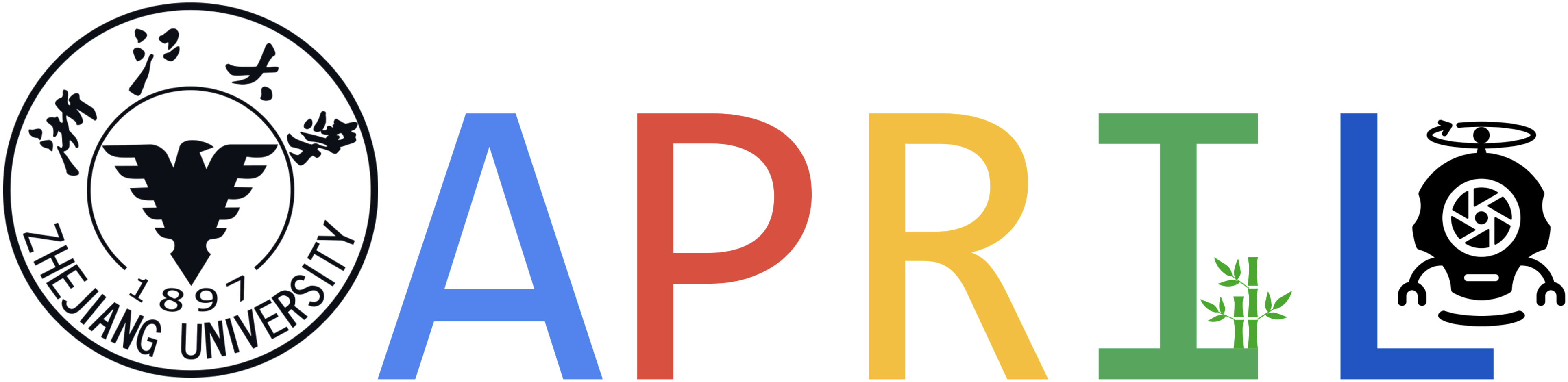}}%
      {}%
  \fi
}
\newcommand{\institutionlogos}{%
  \begin{tikzpicture}[remember picture,overlay]
    \node[
      anchor=north east,
      inner sep=0pt,
      xshift=-2.5cm,
      yshift=-0.78cm
    ] at (current page.north east) {%
      \includegraphics[height=1.52cm]{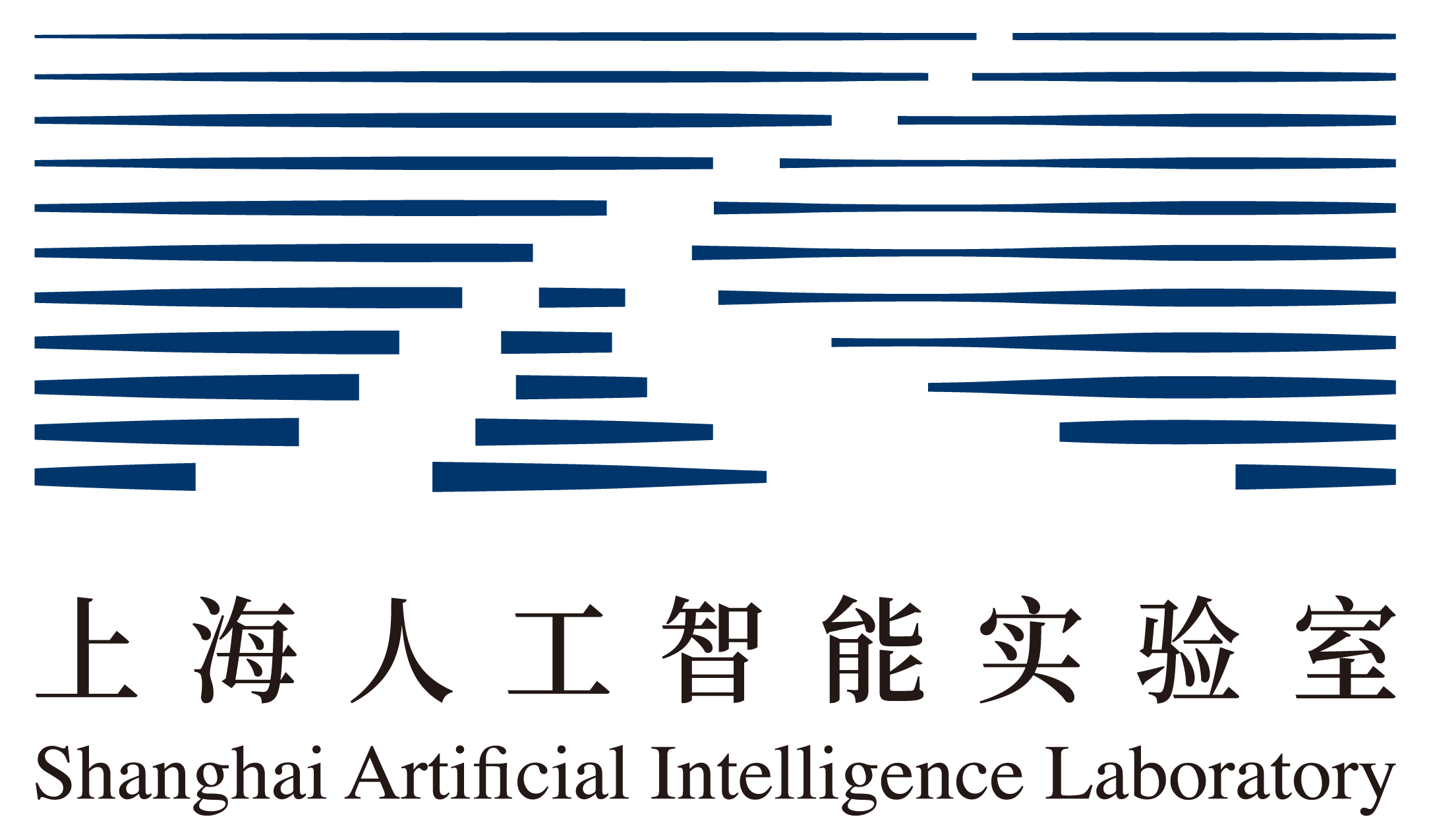}%
      \hspace{0.28cm}%
      \includegraphics[height=1.66cm]{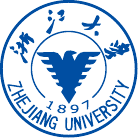}%
      \hspace{0.28cm}%
      \includegraphics[height=1.66cm]{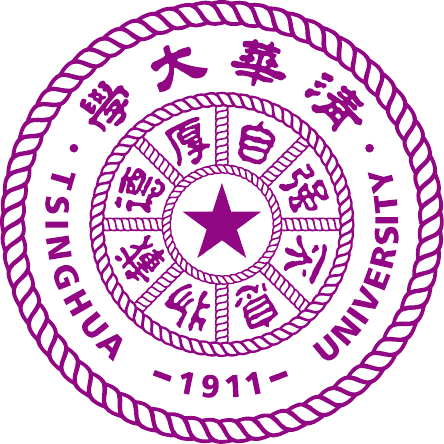}%
    };
  \end{tikzpicture}%
}

\title{%
  \raisebox{-0.15em}{%
    \includegraphics[height=1.3em]{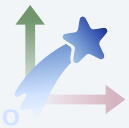}%
  }\hspace{0.15em}%
  VLMs are Native Action Models: Unleashing the Manipulation Capabilities of VLMs with Explicit Action-Space Grounding
}

\author[1,2,*]{Jingkun Yi}
\author[2,3,*]{Shujie Zhang}
\author[3]{Zirui Zhou}
\author[1]{Haoxiao Wang}
\author[1]{Haiyang Sun}
\author[1]{Yuhang Dong}
\author[2]{Xudong Xu}
\author[2]{Xing Shen}
\author[1]{Tao Zhang}
\author[1]{Yong Liu}
\author[1,\dagger]{Jiangning Zhang}

\affiliation[1]{Zhejiang University}
\affiliation[2]{Shanghai Artificial Intelligence Laboratory}
\affiliation[3]{Tsinghua University}

\contribution[*]{Equal contribution}
\contribution[\dagger]{Corresponding author}

\abstract{Pretrained vision-language models (VLMs) already exhibit strong visual, spatial and reasoning capabilities relevant to robotic manipulation. The key challenge in turning pretrained VLMs into generalist manipulation policies is to expose these capabilities through an appropriate action interface. Vision-language-action (VLA) models provide a unified trainable policy, but typically introduce learned action tokens, action heads, or action experts whose semantics must be learned from robot data pretraining. In contrast, modular foundation-model-based manipulation methods can exploit pretrained VLM capabilities for zero-shot manipulation, but rely on external planning and control scaffolds that are not end-to-end trainable as a unified continuous-action policy. In this work, we introduce \textbf{\method{}}, an \textit{\textbf{E}xplicit \textbf{A}ction-\textbf{S}pace \textbf{G}rounding} framework that bridges these two paradigms. By projecting a tool center point~(TCP)-centered action frame into the wrist camera observation and expressing actions in the form of VLM's native textual generation interface, \method{} makes the semantics of a continuous robot action space directly observable in the VLM's perceptual context. This allows pretrained VLMs to directly generate executable robot actions using their existing visual-spatial reasoning capabilities, without robot-action pretraining or additional action modules. Besides, the same action interface supports direct fine-tuning with robot trajectories, allowing the models to continuously improve beyond their manipulation capabilities unleashed by EASG. Overall, \method{} enables zero-shot manipulation and generalization across object positions, categories, and robot embodiments, while supporting effective fine-tuning in the same continuous action space. These results demonstrate that, given an appropriate action interface, pretrained VLMs can themselves serve as \emph{native action models} for robotic manipulation.
}

\covercorrespondence{\email{186368@zju.edu.cn, \{186368, zhangjiangning\}@zju.edu.cn}}
\coversourcecode{https://github.com/General-Robo/EASG}
\coverproject{https://general-robo.github.io/EASG/}

\newif\ifshowtoc
\showtocfalse

\begin{document}

\institutionlogos
\maketitle
\thispagestyle{plain}

\ifshowtoc
  \clearpage
  \setcounter{tocdepth}{2}
  \tableofcontents
  \vspace{1cm}
  \clearpage
\fi

\begin{figure}[htbp]
  \centering
  \includegraphics[width=\textwidth]{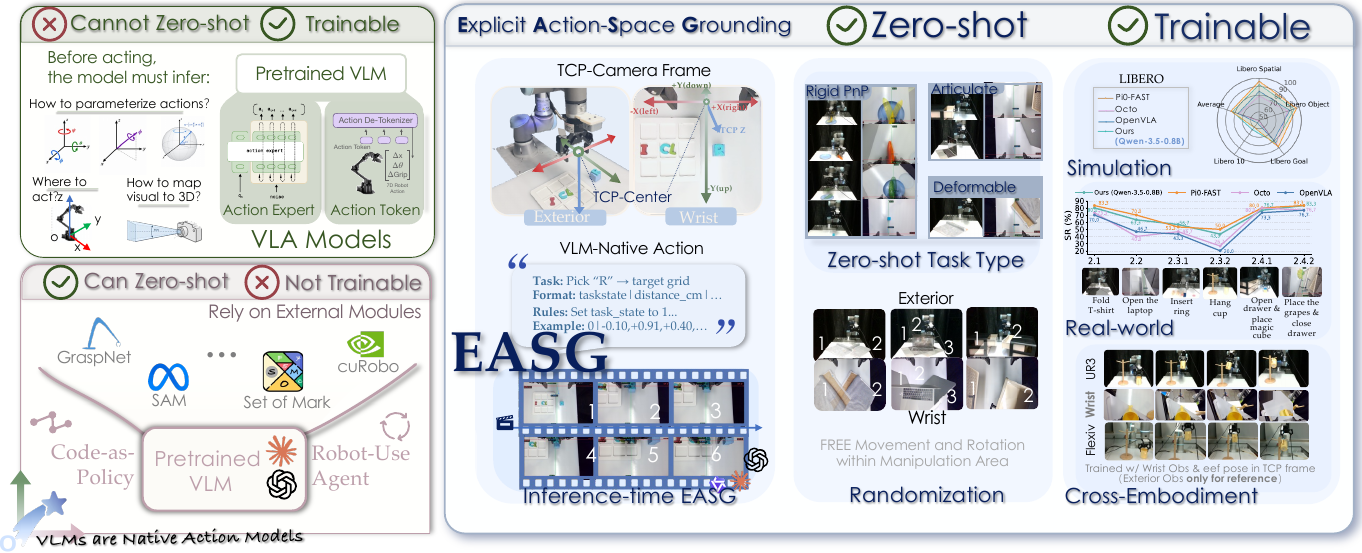}
  \vspace{-0.5em}
  \caption{
    \textbf{VLMs are native action models.} With explicit action space grounding, pretrained VLMs can not only exhibit zero-shot manipulation capabilities without action pretraining, but also be further fine-tuned to improve existing skills and acquire capabilities for manipulation tasks beyond their zero-shot capabilities.
  }
  \label{fig:teaser}
\end{figure}

\section{Introduction}
\label{sec:introduction}

Advances in large-scale multimodal pretraining have equipped vision-language
models (VLMs) with strong visual, spatial, and commonsense reasoning
capabilities~\citep{spatialvlm2024}. Many of these capabilities are directly relevant to robotic
manipulation. Yet turning a pretrained VLM into an executable manipulation
policy requires an appropriate \emph{action interface} that connects its
perceptual and reasoning capabilities to robot control. To make this, such an interface is expected to first specify to the policy how actions should be represented and generated, such as incremental end-effector translation, rotation, and gripper commands. Secondly, it's necessary to clearly state the semantics of the action space to the VLM, including the reference frames and the physical meanings of parameterization. Existing vision-language-action (VLA) models construct such an interface by extending pretrained VLMs with robot-action representations learned from robot trajectories. A common strategy treats robot actions as additional
tokens~\citep{rt22023,openvla2025,fast2025,spatialvla2025}, while other
approaches introduce dedicated action modules, such as flow-matching or
diffusion-based action experts~\citep{pi02024,grootn12025} and continuous action heads trained with regression or parallel decoding
objectives~\citep{openvlaoft2025}. Alongside these architectural advances, VLA
training has increasingly scaled to large and diverse robot
datasets~\citep{openx2024,droid2024,robomind2024,agibotworld2025}. These
approaches provide a unified and trainable action policy and have led to impressive advances in generalist robotic manipulation.

However, the semantics of these robot action spaces are generally not directly available to the pretrained VLM before robot-action training. Robot trajectories must therefore teach the model not only \emph{what action to
take}, but also how that action is defined in the physical world. First, the
model must infer in which coordinate frame an action is expressed and how it
is parameterized, since these conventions are typically only implicit in the
action labels. Second, precise control requires sufficiently accurate metric
geometry. The mapping from 2D visual observations to 3D physical motion is not
uniquely determined without camera geometry~\citep{Hartley2004, zhang2025groundingactionscameraspace, jiang2025knowcameraisviewinvariant}: variations in camera intrinsics
and extrinsics can make visually similar observations correspond to different
robot physical displacements. Third, the policy must learn how embodiment-specific robot states relate to end-effector motion in Cartesian space, including the
relationship between robot configurations, end-effector states, and the
resulting visual changes. Learning these geometric and kinematic
correspondences jointly with manipulation behavior places a substantial burden
on robot data and model capacity. More fundamentally, it prevents the
manipulation-relevant capabilities already present in the pretrained VLM from being directly unleashed to generate executable task-space actions. Besides, extensive robot-action training may
further alter some of the visual-semantic and reasoning capabilities inherited
from the pretrained VLM~\citep{driess2025knowledge,hancock2026actions}.

An alternative line of work exposes pretrained foundation-model capabilities
through explicit intermediate representations instead of learning the entire action interface from robot trajectories. These \textbf{modular foundation-model-based} manipulation methods generate policy programs~\citep{codeaspolicies2022},
spatial constraints over object parts~\citep{copa2024}, composable 3D value
maps~\citep{voxposer2023} or relational keypoint
constraints~\citep{rekep2025}, which are subsequently converted into robot
motion by downstream planners~\citep{kirillov2023segment,yang2023setofmark}, optimizers~\citep{curobo_v2} or controllers~\citep{fang2020graspnet}. By formulating manipulation problems in representations that pretrained foundation models can
already reason about, these methods can perform a range of tasks zero-shot without robot-action pretraining. However, the mapping depends on intermediate representations and external planning or execution modules without a unified action policy. Consequently,
the same VLM policy cannot be trained end-to-end in the continuous action space with robot trajectories through the interface that enables its zero-shot behavior.

The two paradigms therefore expose a complementary trade-off. VLA models
provide a unified and trainable action policy, but require the action interface
to be learned from robot data before the model can act. Modular
foundation-model methods can directly exploit pretrained capabilities for zero-shot manipulation, but externalize action generation across separate
representations, planners, and controllers. This raises a natural question: \textit{Given an appropriate action interface, can a pretrained VLM serve directly as a unified manipulation policy that acts zero-shot
and can be further improved through robot-data fine-tuning?}

To answer this question, we introduce \methodlit{}, an
\textit{\textbf{E}xplicit \textbf{A}ction-\textbf{S}pace \textbf{G}rounding}
framework that provides a unified action interface between pretrained VLM reasoning
and continuous robot control. \methodlit{} explicitly grounds the semantics of
a continuous robot action space in the VLM's perceptual context by projecting
a tool center point (TCP)-centered action frame into first-person observations in the wrist view and providing metric information for action execution. Actions are expressed through the VLM's native textual generation interface, so the pretrained model can directly map its existing visual-spatial reasoning to continuous robot
actions without adapting to a new robot-specific output modality.
Together, these designs allow pretrained VLMs to generate executable actions
without robot-action pretraining or auxiliary learned planning modules.
Crucially, the same action space and generation interface directly support robot-data fine-tuning, allowing the VLM to continuously improve the
manipulation capabilities initially unleashed by \methodlit{}. 
To ensure reliable action generation through this native interface, we
further introduce grammar-constrained decoding, which restricts autoregressive
generation to valid action formats while preserving the pretrained VLM's
original vocabulary and generation process.
Because the pretrained VLM itself generates robot actions through the same native
generation interface both before and after fine-tuning, we refer to it as a
\emph{native action model}.
To summarize, our key contributions are as follows:

\begin{itemize}

\item We introduce \methodlit{}, an \textit{Explicit Action-Space Grounding}
framework that makes robot-action semantics directly accessible to
pretrained VLMs while preserving their native textual generation interface,
enabling them to act without large-scale robot-action pretraining.

\item We show that the same action interface supports both zero-shot manipulation and robot-data fine-tuning.

\item  We empirically demonstrate the capability of VLM-native manipulation:
a 27B-scale VLM with EASG achieves comparable zero-shot performance with
fine-tuned VLA models, while a 0.8B-scale VLM fine-tuned with EASG achieves
competitive performance with several VLA baselines.

\end{itemize}

\section{Related Work}

\paragraph{Vision-Language-Action (VLA) Models.} Existing VLA models have explored discrete action tokens~\citep{rt22023,openvla2025,belkhale2024rth,fast2025,zhen2024_3dvla,spatialvla2025} or additional action modules~\citep{pi02024,grootn12025,openvlaoft2025} for better robotic action representation and generation. With these architectural advances, VLA pretraining has scaled to increasingly large and diverse robot datasets~\citep{brohan2023rt1,openx2024,droid2024,robomind2024,agibotworld2025}. More recently, video or world-model-based policies also try to leverage pretrained video generators to provide future visual dynamic priors for action generation~\citep{cosmospolicy2026, dreamzero2026}. Despite these advances, the prevailing action-pretraining paradigm remains highly data-intensive, since it implicitly requires models to jointly infer action coordinates, depth information and inverse dynamics from robot data. We provide further discussion in Appendix~\ref{app:related-work-vla}.

\paragraph{Modular Foundation-Model-Based Manipulation.}
Modular foundation-model-based manipulation methods leverage pretrained
foundation models for high-level perception and reasoning~\citep{huang2022zeroshotplanner, zeng2022socratic}, while mediating
robot action through explicit intermediate representations and downstream
planning or control modules~\citep{singh2023progprompt}.
Code as Policies~\citep{codeaspolicies2022} generates executable programs that
compose predefined perception and control APIs, while subsequent methods use
structured representations such as 3D affordance maps in
VoxPoser~\citep{voxposer2023}, spatial end-effector constraints in
CoPa~\citep{copa2024}, and relational keypoint constraints in
ReKep~\citep{rekep2025}.
These methods enable zero-shot manipulation without robot-action pretraining,
and recent \textbf{Robot-Use-Agent} approaches further expose perception, planning, and control as callable tools or autonomously refine robot policies through execution
feedback~\citep{capx2026,via2026,enpire2026,aspire2026}.
However, their perception-to-action mapping remains distributed across the
foundation model and external execution modules rather than forming a unified action policy within the model itself, preventing end-to-end optimization of the complete perception-to-action pipeline.
Further discussion is provided in Appendix~\ref{app:related-work-code-as-policy}.

\paragraph{3D Visuomotor Manipulation.} A complementary line of work incorporates explicit 3D geometry into visuomotor policy learning~\citep{ke2025diffuseractor}. PerAct~\citep{peract} predicts 6-DoF actions from voxelized RGB-D scenes, while RVT~\citep{rvt} and RVT-2~\citep{rvt2} use virtual multi-view projections and coarse-to-fine prediction for efficient spatial reasoning. In contrast, \method{} does not reconstruct the full 3D scene, but dynamically projects the $z$-axis of the TCP-centered frame toward the manipulation target. This action-centric grounding with depth information enables pretrained VLMs to directly reason about 3D motion for zero-shot manipulation and subsequent fine-tuning.
\section{Method}
\label{sec:method}

\begin{figure}[htbp]
    \centering
    \includegraphics[width=\textwidth]{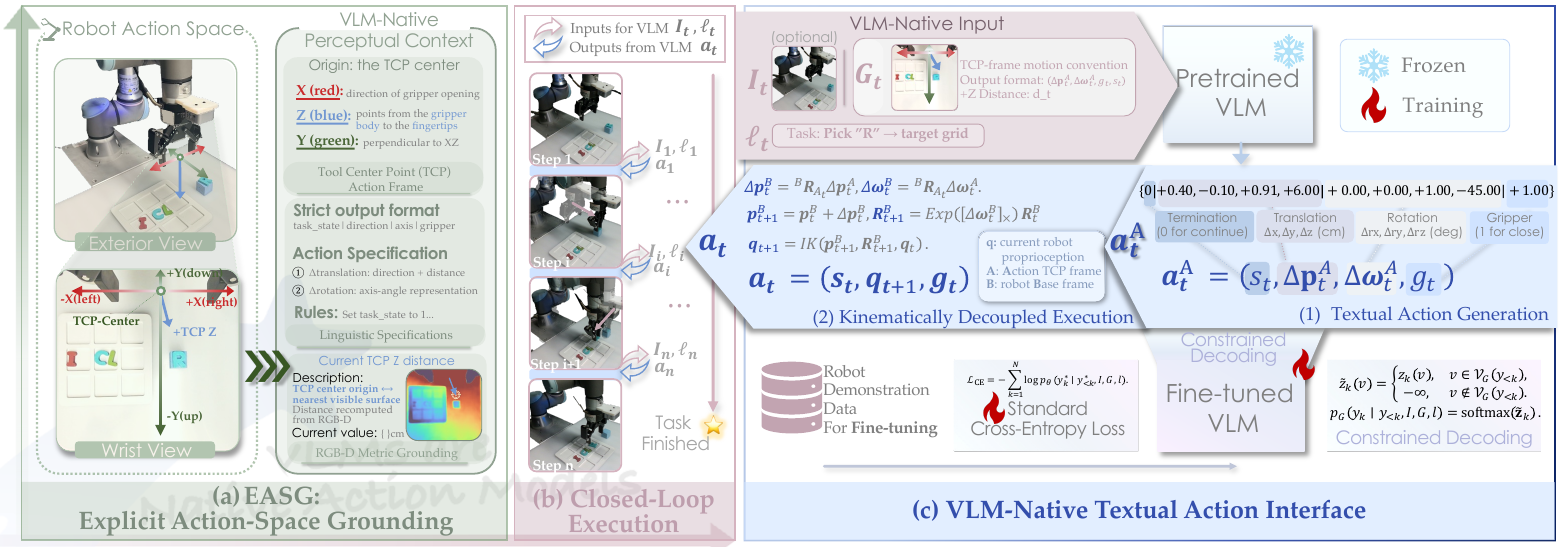}
    \caption{
    \textbf{Overview of \methodlit{}.}
    \methodlit{} explicitly grounds robot-action semantics in the
    VLM's perceptual context and expresses actions through its native textual
    generation interface, enabling both zero-shot manipulation and direct
    robot-data fine-tuning. Actions are generated in the TCP-centered frame(Action frame) and converted into low-level joint commands.
    }
    \label{fig:method-overview}
\end{figure}

\textbf{Explicit Action-Space Grounding (EASG)} makes the physical semantics
of a robot action space directly observable in the VLM's
perceptual context. We define the action-space grounding $G_t$ as a structured
representation that exposes the semantics of the continuous action
space to the VLM:
\begin{equation}
G_t = (V_t, M_t, P_t),
\end{equation}
where $V_t$ denotes the visual projection of the TCP-centered action
frame, $M_t$ provides metric distance information by estimating the
distance from the TCP center to the nearest visible surface along the
TCP approach direction, and $P_t$ specifies the linguistic semantics of
the action parameterization and output format.
At each control step $t$, the policy receives visual observations $I_t$, a
language instruction $l$, and action-space grounding $G_t$, and predicts
\begin{equation}
    a_t =
    \left(
        \Delta \mathbf{p}^{A}_t,\,
        \Delta \boldsymbol{\omega}^{A}_t,\,
        g_t,\,
        s_t
    \right),
    \label{eq:action}
\end{equation}
where $\Delta \mathbf{p}^{A}_t\in\mathbb{R}^{3}$ is an incremental
translation, $\Delta \boldsymbol{\omega}^{A}_t\in\mathbb{R}^{3}$ is an
incremental axis--angle rotation vector, both expressed in the grounded
action frame $\mathcal{F}_{A_t}$, $g_t$ is the gripper command, and $s_t$
indicates task completion.
The action is executed, the environment is re-observed, and the process
repeats in closed loop.

\subsection{Explicit Action-Space Grounding}
\label{sec:easg}

EASG explicitly grounds the origin, axes, parameterization, and metric scale of the action space in the current observation, rather than requiring these semantics to be inferred implicitly from robot demonstrations.
At each control step, we define a local action frame
$\mathcal{F}_{A_t}$ whose origin is anchored at the current Tool Center Point
(TCP). Its axes define the directions in which the VLM expresses incremental
translation and rotation.
Given calibrated camera geometry and the current robot state, the action frame
is represented in the wrist-camera coordinate system as
\begin{equation}
    {}^{C}\mathbf{T}_{A_t}
    =
    \begin{bmatrix}
        {}^{C}\mathbf{R}_{A_t} &
        {}^{C}\mathbf{p}_{A_t}\\
        \mathbf{0}^{\top} & 1
    \end{bmatrix}.
    \label{eq:camera_action_frame}
\end{equation}
Let $\mathbf{K}$ denote the camera intrinsic matrix and $\pi(\cdot)$ the
perspective projection operator.
We project the action-frame origin and axes into the wrist-camera image:
\begin{equation}
\begin{aligned}
    \mathbf{u}_0
    &=
    \pi\!\left(
        \mathbf{K}\,{}^{C}\mathbf{p}_{A_t}
    \right),
    \\
    \mathbf{u}_j
    &=
    \pi\!\left[
        \mathbf{K}
        \left(
            {}^{C}\mathbf{p}_{A_t}
            +
            \ell\,{}^{C}\mathbf{R}_{A_t}\mathbf{e}_j
        \right)
    \right],
    \qquad j\in\{x,y,z\},
\end{aligned}
\label{eq:action_projection}
\end{equation}
where $\mathbf{e}_j$ is a canonical action direction and $\ell$ controls
the visualized axis length.
The accompanying prompt specifies axis directions, translation units,
rotation parameterization, and gripper semantics.
The VLM therefore both observes the action coordinate system in its visual
context and receives an explicit interpretation of its numerical outputs.
To provide metric scale, we use aligned depth to estimate scene distance along
the action-frame approach direction.
For the ray
\begin{equation}
    \mathbf{r}(d)
    =
    {}^{C}\mathbf{p}_{A_t}
    +
    d\,{}^{C}\mathbf{R}_{A_t}\mathbf{e}_z,
    \label{eq:z_ray}
\end{equation}
we estimate the distance $d_t$ to the nearest visible surface from the RGB-D
observation.
This supplies metric 3D information directly, rather than requiring the VLM
to infer precise physical distance from RGB appearance alone.
The VLM predicts motion only in the grounded action frame
$\mathcal{F}_{A_t}$; robot-specific coordinate transforms and kinematics are
handled deterministically.
Let $\mathcal{F}_{B}$ denote the robot base frame and
${}^{B}\mathbf{R}_{A_t}$ the orientation of the action frame expressed in
$\mathcal{F}_{B}$.
The predicted translation and rotation increments are first transformed into
the robot base frame:
\begin{equation}
\Delta\mathbf{p}^{B}_t
=
{}^{B}\mathbf{R}_{A_t}
\Delta\mathbf{p}^{A}_t,
\qquad
\Delta\boldsymbol{\omega}^{B}_t
=
{}^{B}\mathbf{R}_{A_t}
\Delta\boldsymbol{\omega}^{A}_t.
\label{eq:action_to_base}
\end{equation}
%
Let $\mathbf{p}^{B}_t$ and $\mathbf{R}^{B}_t$ denote the current controlled
end-effector position and orientation in the robot base frame.
The corresponding target pose is
\begin{equation}
\mathbf{p}^{B,*}_t
=
\mathbf{p}^{B}_t
+
\Delta\mathbf{p}^{B}_t,
\qquad
\mathbf{R}^{B,*}_t
=
\operatorname{Exp}
\left(
[\Delta\boldsymbol{\omega}^{B}_t]_{\times}
\right)
\mathbf{R}^{B}_t .
\label{eq:base_target_pose}
\end{equation}
%
We denote the resulting target end-effector pose in the base frame by
$\mathbf{T}^{B,*}_t$ and convert it into robot-specific commands through
inverse kinematics and low-level control:
\begin{equation}
    \mathbf{q}^{*}_t
    =
    \operatorname{IK}
    \left(
        \mathbf{T}^{B,*}_t;
        \mathbf{q}_t
    \right).
    \label{eq:ik}
\end{equation}
Thus, the VLM reasons in a common grounded action space, while deterministic
geometry and embodiment-specific kinematics realize the predicted motion on
each robot.

\subsection{VLM-Native Textual Action Learning}
\label{sec:textual_action_learning}

We serialize continuous robot actions using only tokens from the pretrained
VLM's native textual vocabulary, without introducing robot-specific action
tokens or a separate continuous-action predictor.
First, we parameterize translation by a unit direction
$\mathbf{v}^{A}_t\in\mathbb{R}^{3}$ and a nonnegative distance $d_t$,
and rotation by a unit axis
$\mathbf{r}^{A}_t\in\mathbb{R}^{3}$ and a signed angle $\theta_t$:
\begin{equation}
    \Delta\mathbf{p}^{A}_t
    =
    d_t\,\mathbf{v}^{A}_t,
    \qquad
    \Delta\boldsymbol{\omega}^{A}_t
    =
    \theta_t\,\mathbf{r}^{A}_t,
    \label{eq:action_parameterization}
\end{equation}
where
$\|\mathbf{v}^{A}_t\|_2=1$ for nonzero translation and
$\|\mathbf{r}^{A}_t\|_2=1$ for nonzero rotation.
Both directions are expressed in the grounded action frame
$\mathcal{F}_{A_t}$.
We represent $d_t$ in centimeters and $\theta_t$ in degrees.
In practice, the VLM
may generate vectors that do not exactly satisfy the unit-norm constraint.
We therefore project the generated vectors onto the unit sphere before
executing the action.
The resulting textual action follows
\begin{equation}
    y_t =
    \texttt{
    state
    \;|\;
    $v_x,v_y,v_z,d$
    \;|\;
    $r_x,r_y,r_z,\theta$
    \;|\;
    gripper
    },
    \label{eq:text_action}
\end{equation}
where every continuous scalar is emitted with an explicit \texttt{+} or
\texttt{-} sign and exactly two decimal places.
For example,
\begin{center}
\small
\texttt{
0|+0.40,-0.10,+0.91,+6.00|
+0.00,+0.00,+1.00,-45.00|+1.00
}.
\end{center}
The first field indicates task completion, while the final field specifies the
desired gripper opening.
Because all symbols already belong to the pretrained language vocabulary, the
original autoregressive language decoder itself serves as the action decoder.
The same textual action representation is used for both zero-shot execution
and subsequent robot-data fine-tuning.
Given a demonstrated action sequence
$y^*=(y_1^*,\ldots,y_N^*)$, we therefore fine-tune the VLM directly with
standard causal language-model cross-entropy:
\begin{equation}
    \mathcal{L}_{\mathrm{CE}}
    =
    -\sum_{k=1}^{N}
    \log
    p_{\theta}
    \left(
        y_k^*
        \mid
        y_{<k}^*, I, G, l
    \right).
    \label{eq:ce_loss}
\end{equation}
Fine-tuning  improves the same autoregressive policy that already
supports zero-shot manipulation, without changing its action interface.
After fine-tuning, directly decoding actions as free-form language may produce invalid action formats, including malformed numeric values, missing fields, incorrect separators, or extraneous text. We therefore enforce a deterministic action grammar to constrain decoding and ensure valid action generation~\citep{geng-etal-2023-grammar}.
Let $\mathcal{V}_{G}(y_{<k})$ denote the set of tokens valid after prefix
$y_{<k}$.
For the original VLM logit $z_k(v)$, we use
\begin{equation}
    \widetilde{z}_k(v)
    =
    \begin{cases}
        z_k(v),
        & v\in\mathcal{V}_{G}(y_{<k}),\\
        -\infty,
        & \text{otherwise}.
    \end{cases}
    \label{eq:constrained_logits}
\end{equation}
Generation then proceeds autoregressively from
$\operatorname{softmax}(\widetilde{\mathbf{z}}_k)$.
The grammar constrains only command syntax---including field order,
separators, signs, and numerical precision---while the numerical action itself
remains predicted by the VLM.
Further details are provided in Appendix~\ref{app:constrained-decoding}.

\section{Experiments}
\label{sec:experiments}

\begin{wrapfigure}{l}{0.36\columnwidth}
\centering

\includegraphics[width=\linewidth]{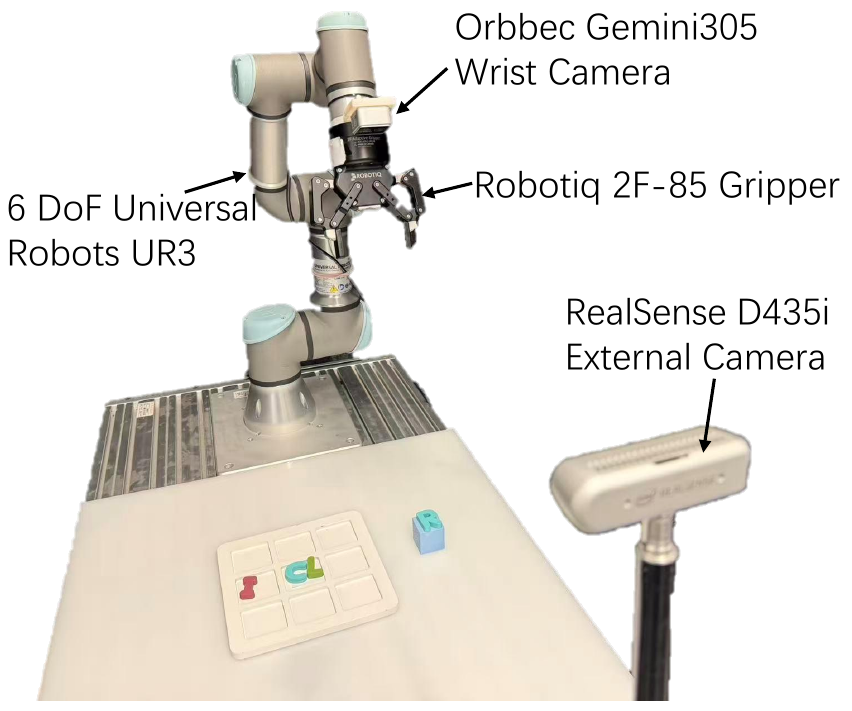}

\caption{\textbf{Real-world robot setup.}}
\label{fig:real-setup}

\end{wrapfigure}

\paragraph{Experimental Setup.}

To study both the native zero-shot capability and the extensibility of VLM-based manipulation, we conduct real-world experiments on a 6-DoF UR3 robot with a wrist-mounted RGB-D camera and an external camera. We construct a task suite spanning rigid, articulated, deformable, precise, and long-horizon manipulation, and organize it into two groups of different difficulty. Task Group 1 contains relatively simple tasks, whereas Task Group 2 includes more challenging tasks requiring greater precision, more complex interactions, or longer-horizon execution. We find that pretrained VLMs with EASG exhibit non-trivial zero-shot capability on Task Group 1 but generally fail on Task Group 2. Accordingly, we use Task Group 1 to characterize native VLM manipulation and Task Group 2 to evaluate capability extension through robot-data fine-tuning.

\vspace{2.0em}
\begin{table*}[htbp]
\centering

\begin{minipage}[t]{0.385\textwidth}
\centering

\textbf{Task Group 1: Zero-shot evaluation tasks.}

\vspace{0.35em}

\footnotesize
\setlength{\tabcolsep}{3.5pt}
\renewcommand{\arraystretch}{0.85}

\begin{tabularx}{\linewidth}{
    @{}
    >{\centering\arraybackslash}p{0.21\linewidth}
    >{\centering\arraybackslash}p{0.14\linewidth}
    X
    @{}
}
\toprule
\textbf{Type}
& \textbf{Task}
& \textbf{Task description} \\
\midrule

\multirow{3}{*}{\raisebox{-1.5ex}{Rigid}}
& 1.1.1
& Grasp the cucumber \\

\cmidrule(lr){2-3}
&
1.1.2
& Pick up the block \\

\cmidrule(lr){2-3}
&
1.1.3
& Pick up the banana and put it on the plate \\

\cmidrule(lr){1-3}

\multirow{2}{*}{Articulated}
& 1.2.1
& Close the laptop \\

\cmidrule(lr){2-3}
&
1.2.2
& Push the drawer closed \\

\cmidrule(lr){1-3}

Deformable
& 1.3
& Fold the towel \\

\bottomrule
\end{tabularx}

\end{minipage}
\hfill
\begin{minipage}[t]{0.55\textwidth}
\centering

\textbf{Task Group 2: Capability extension tasks.}

\vspace{0.35em}

\footnotesize
\setlength{\tabcolsep}{4.5pt}
\renewcommand{\arraystretch}{0.85}

\begin{tabularx}{\linewidth}{
    @{}
    >{\centering\arraybackslash}p{0.18\linewidth}
    >{\centering\arraybackslash}p{0.11\linewidth}
    X
    @{}
}
\toprule
\textbf{Type}
& \textbf{Task}
& \textbf{Task description} \\
\midrule

Deformable
& 2.1
& Fold the T-shirt in half. \\

\cmidrule(lr){1-3}

Articulated
& 2.2
& Open the laptop. \\

\cmidrule(lr){1-3}

\multirow{2}{*}{Precise}
& 2.3.1
& Pick up the ring and put it on one of the pegs. \\

\cmidrule(lr){2-3}
&
2.3.2
& Hang the cup on the peg. \\

\cmidrule(lr){1-3}

\multirow{2}{*}{Long-horizon}
& 2.4.1
& Open the drawer, then put the Rubik's Cube in the drawer. \\

\cmidrule(lr){2-3}
&
2.4.2
& Put the grapes in the drawer, then close the drawer. \\

\bottomrule
\end{tabularx}

\end{minipage}

\caption{
\textbf{Real-world task setups.}
See Appendix~\ref{app:real-zero-exp} for the full object group we included.
}
\label{fig:real-tasks}

\end{table*}

\par\WFclear\FloatBarrier

\subsection{Zero-Shot VLM Manipulation via Action-Space Grounding}
\label{sec:exp-zero-shot-bridge}

\begin{expquestion}{VLMFrost}{
    \textit{Q1. Can pretrained VLMs perform zero-shot manipulation with explicit action-space grounding?}
}
\end{expquestion}

\begin{table}[htbp]
\centering

\caption{
\textbf{Robot data efficiency comparison between EASG and fine-tuned VLAs.}
Ours uses no task-specific demonstrations. VLA baselines are
fine-tuned with 10/30/50 demonstrations per task.
}
\label{tab:zeroshot-vla-finetune}
\scriptsize
\setlength{\tabcolsep}{1.2pt}
\renewcommand{\arraystretch}{0.88}

\begin{tabularx}{\linewidth}{
    >{\centering\arraybackslash}p{0.055\linewidth}
    >{\columncolor{VLMFrost!35}}Y
    >{\columncolor{VLMCream!15}}Y
    >{\columncolor{VLMFrost!35}}Y
    >{\columncolor{VLMSky!55}}Y
    >{\columncolor{VLMCream!15}}Y
    >{\columncolor{VLMFrost!35}}Y
    >{\columncolor{VLMSky!55}}Y
    >{\columncolor{VLMCream!15}}Y
    >{\columncolor{VLMFrost!35}}Y
    >{\columncolor{VLMSky!55}}Y
    >{\columncolor{VLMCream!15}}Y
    >{\columncolor{VLMFrost!35}}Y
    >{\columncolor{VLMSky!55}}Y
}
\toprule

\multirow{2}{*}{\textbf{Task}}
&
\multicolumn{1}{c}{\textbf{Ours}}
&
\multicolumn{3}{c}{\textbf{$\boldsymbol{\pi}_{0.5}$}}
&
\multicolumn{3}{c}{\textbf{$\boldsymbol{\pi}_{0}$-FAST}}
&
\multicolumn{3}{c}{\textbf{GR00T N1.6}}
&
\multicolumn{3}{c}{\textbf{OpenVLA-OFT}}
\\[-1pt]

\cmidrule(lr){3-5}
\cmidrule(lr){6-8}
\cmidrule(lr){9-11}
\cmidrule(lr){12-14}

&
\textit{Zero}
&
10 & 30 & 50
&
10 & 30 & 50
&
10 & 30 & 50
&
10 & 30 & 50
\\

\midrule

1.1.1 & 93.3
& 70.0 & 100.0 & 100.0
& 53.3 & 90.0 & 100.0
& 66.7 & 100.0 & 100.0
& 66.7 & 100.0 & 100.0 \\

1.1.2 & 90.0
& 66.7 & 100.0 & 100.0
& 53.3 & 86.7 & 100.0
& 60.0 & 100.0 & 100.0
& 66.7 & 93.3 & 100.0 \\

1.1.3 & 66.7
& 30.0 & 80.0 & 100.0
& 0.0 & 46.7 & 73.3
& 10.0 & 70.0 & 100.0
& 20.0 & 53.3 & 93.3 \\

1.2.1 & 100.0
& 50.0 & 80.0 & 100.0
& 46.7 & 53.3 & 100.0
& 50.0 & 53.3 & 100.0
& 53.3 & 66.7 & 100.0 \\

1.2.2 & 93.3
& 46.7 & 66.7 & 100.0
& 30.0 & 53.3 & 100.0
& 46.7 & 50.0 & 100.0
& 40.0 & 70.0 & 100.0 \\

1.3 & 33.3
& 0.0 & 30.0 & 80.0
& 0.0 & 10.0 & 40.0
& 0.0 & 33.3 & 73.3
& 0.0 & 13.3 & 40.0 \\

\midrule

\textbf{Avg.} & 79.4
& \textbf{43.9} & \textbf{76.1}  & \textbf{96.7}
& 30.6 & 56.7 & 85.6
& 38.9 & 67.8 & 95.6
& 41.1 & 66.1 & 88.9 \\

\bottomrule
\end{tabularx}

\end{table}

\paragraph{Zero-shot Capability.} To probe the native zero-shot manipulation capabilities of pretrained VLMs with \method, we evaluate Qwen-3.8-27B~\citep{qiu2026qwen38} on Task Group 1, as summarized in Table~\ref{fig:real-tasks}. For each task, we conduct three independent evaluation rounds with 10 trials per round, resulting in 30 trials in total. We report the mean success rate and standard deviation across the three evaluation rounds. The linguistic action-space grounding used in our experiments is provided in Appendix~\ref{app:prompts}.


\paragraph{Robot Data Needed to Match \method{} Zero-Shot Performance.} General-purpose VLAs typically require a small amount of task-specific data to adapt to a new embodiment and domain before executing newly specified tasks. We therefore fine-tune $\pi_{0.5}$~\citep{pi05_2025}, $\pi_0$-FAST~\citep{fast2025}, GR00T N1.6~\citep{grootn12025}, and OpenVLA-OFT~\citep{openvlaoft2025} using 10/20/30/40/50 demonstrations per task. We find that approximately 30 demonstrations are sufficient for fine-tuned VLA policies to match the zero-shot success rate of \method{}. Table~\ref{tab:zeroshot-vla-finetune} reports representative results at 10/30/50 demonstrations. The \textbf{full five-level comparison} is in Table~\ref{tab:zeroshot-vla-finetune-full}.

\begin{wrapfigure}[10]{r}{0.45\linewidth}
    \centering

    \includegraphics[width=0.85\linewidth]{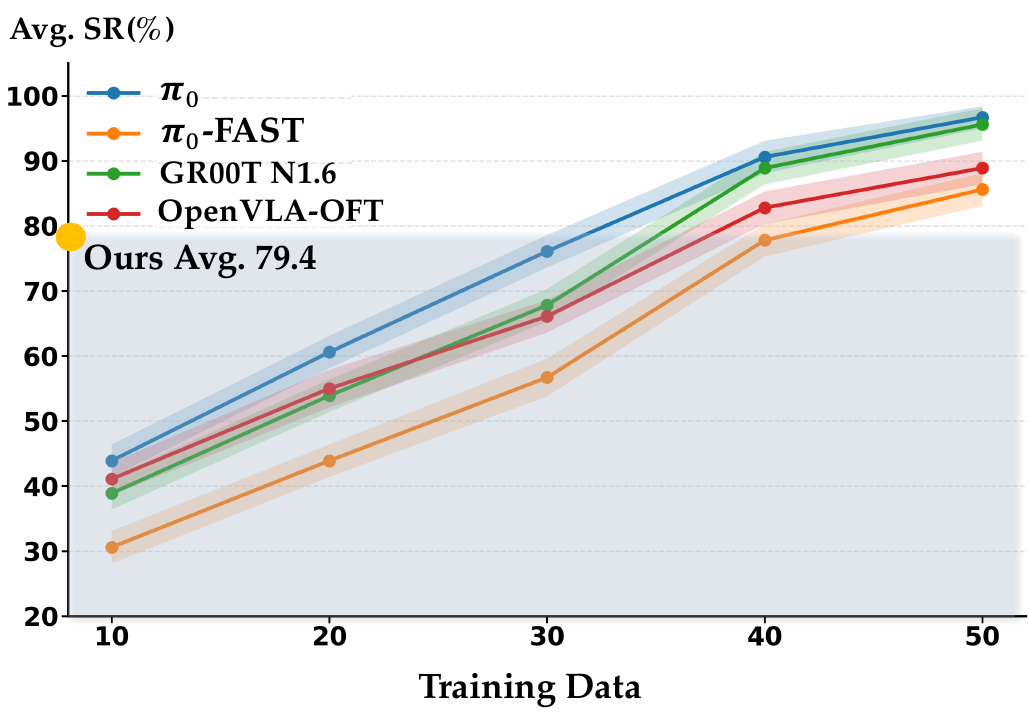}

    \caption{
    \textbf{Results visualization.}
    }
    
    \label{fig:data-needed-match}
\end{wrapfigure}

\paragraph{Zero-Shot Generalization across Objects and Positions.} 
Since \method{} leverages the visual understanding acquired during VLM pretraining, we expect it to be naturally robust to variations in object positions, poses and color. We deploy three randomized configurations in Task Group 1, including pose, position and color. 
In Table~\ref{tab:zeroshot-randomization}, we find \method{} largely preserves zero-shot performance of VLMs, whereas VLA baselines fine-tuned with 50 demonstrations exhibit
a clear performance drop despite deliberate randomization of object poses and positions during VLA demonstration collection. For detailed setups and data analysis, see Appendix~\ref{app:real-zero-random}.

\vspace{3.0em}





\begin{table}[htbp]

\centering
\caption{
\textbf{Generalization under object and position randomization.}
Randomized results are reported as mean $\pm$ Std. over three randomized configurations.
}
\label{tab:zeroshot-randomization}
\footnotesize
\setlength{\tabcolsep}{2pt}
\renewcommand{\arraystretch}{0.9}

\begin{tabular}{
@{}
>{\raggedright\arraybackslash}p{0.16\linewidth}
>{\centering\arraybackslash}p{0.13\linewidth}
>{\centering\arraybackslash}p{0.13\linewidth}
>{\centering\arraybackslash}p{0.13\linewidth}
>{\centering\arraybackslash}p{0.18\linewidth}
>{\centering\arraybackslash}p{0.21\linewidth}
@{}
}
\toprule
&
\textbf{Ours}(Zero)
&
\textbf{$\pi_{0.5}$}(50)
&
\textbf{$\pi_{0}$-FAST}(50)
&
\textbf{GR00T N1.6}(50)
&
\textbf{OpenVLA-OFT}(50)
\\
\midrule

\textbf{Randomized}
& 77.3$\pm$5.9
& 40.2$\pm$8.7
& 25.5$\pm$9.5
& 37.8$\pm$10.6
& 28.1$\pm$10.9
\\

\textbf{Original}
&
79.4$\pm$2.5&
96.7$\pm$1.7&
85.6$\pm$2.5&
95.6$\pm$2.5&
88.9$\pm$2.5
\\
\bottomrule
\end{tabular}

\end{table}


\begin{expfinding}{VLMFrost}
\textbf{\textit{Findings.}} Explicit action-space grounding turns pretrained VLMs into zero-shot manipulation models with strong position, pose and color generalization.
\end{expfinding}

\subsection{Improving VLM-Native Manipulation through Post-Training}
\label{sec:exp-post-training}

\begin{expquestion}{VLMSky}{
    \textit{Q2. Can VLMs be fine-tuned beyond their zero-shot manipulation capabilities?}
}
\end{expquestion}

\paragraph{Fine-tuning VLMs in the Real World.} We organize several more challenge tasks that cannot be completed by our EASG framework zero-shot
, as depicted in Task Group 2, which contains manipulating deformable, articulated tasks, precise operation, and long-horizon tasks. We fine-tune the qwen-3.8-27B following the textual action learning procedure described in Sec.~\ref{sec:textual_action_learning} with Lora~\citep{hu2021lora}. And we want to determine if it can extend its capability through finetuning. We include $\pi_0$-FAST, OpenVLA\citep{openvla2025}, Octo\citep{octo2024} as ours baselines for comparison. 



\begin{table}[H]

\centering

\caption{
\textbf{Results on harder real-world tasks.}
\textbf{FT} denotes fine-tuning with teleoperated data.
Gray cells indicate unavailable results, as pretrained VLA models generally need additional adaptation.
}

\label{tab:zero-finetune-comparison}

\scriptsize
\setlength{\tabcolsep}{1.2pt}
\renewcommand{\arraystretch}{0.88}

\newcolumntype{Z}{
    >{\columncolor{VLMCream}\centering\arraybackslash}X
}

\newcolumntype{F}{
    >{\columncolor{VLMSky}\centering\arraybackslash}X
}

\newcolumntype{A}{
    >{\columncolor{VLMSky!140}\centering\arraybackslash}X
}

\begin{tabularx}{\linewidth}{
    @{}
    >{\raggedright\arraybackslash}p{0.12\linewidth}
    *{6}{ZF}
    *{6}{F}
    A
    @{}
}

\toprule

\multirow{2}{*}{\textbf{Method}}
&
\multicolumn{2}{c}{\textbf{1.1.1}}
&
\multicolumn{2}{c}{\textbf{1.1.2}}
&
\multicolumn{2}{c}{\textbf{1.1.3}}
&
\multicolumn{2}{c}{\textbf{1.2.1}}
&
\multicolumn{2}{c}{\textbf{1.2.2}}
&
\multicolumn{2}{c}{\textbf{1.3}}
&
\multicolumn{1}{c}{\textbf{2.1}}
&
\multicolumn{1}{c}{\textbf{2.2}}
&
\multicolumn{1}{c}{\textbf{2.3.1}}
&
\multicolumn{1}{c}{\textbf{2.3.2}}
&
\multicolumn{1}{c}{\textbf{2.4.1}}
&
\multicolumn{1}{c}{\textbf{2.4.2}}
&
\multicolumn{1}{c}{\textbf{Avg.}}
\\[-1pt]

\cmidrule(lr){2-3}
\cmidrule(lr){4-5}
\cmidrule(lr){6-7}
\cmidrule(lr){8-9}
\cmidrule(lr){10-11}
\cmidrule(lr){12-13}
\cmidrule(lr){14-14}
\cmidrule(lr){15-15}
\cmidrule(lr){16-16}
\cmidrule(lr){17-17}
\cmidrule(lr){18-18}
\cmidrule(lr){19-19}
\cmidrule(lr){20-20}

&
\textit{Zero} & \textit{FT}
&
\textit{Zero} & \textit{FT}
&
\textit{Zero} & \textit{FT}
&
\textit{Zero} & \textit{FT}
&
\textit{Zero} & \textit{FT}
&
\textit{Zero} & \textit{FT}
&
\textit{FT}
&
\textit{FT}
&
\textit{FT}
&
\textit{FT}
&
\textit{FT}
&
\textit{FT}
&
\textit{FT}
\\

\midrule

\textbf{Ours}
& 93.3 & 100.0
& 90.0 & 100.0
& 66.7 & 70.0
& 100.0 & 100.0
& 93.3 & 100.0
& 33.3 & 33.3
& 80.0
& 63.3
& 56.7
& 43.3
& 76.7
& 83.3
& 75.5
\\

$\pi_{0}$-FAST
& \cellcolor{VLANoResult} -- & 100.0
& \cellcolor{VLANoResult} -- & 100.0
& \cellcolor{VLANoResult} -- & 73.3
& \cellcolor{VLANoResult} -- & 100.0
& \cellcolor{VLANoResult} -- & 100.0
& \cellcolor{VLANoResult} -- & 40.0
& 83.3
& 70.0
& 53.3
& 50.0
& 80.0
& 83.3
& 77.8
\\

OpenVLA
& \cellcolor{VLANoResult} -- & 100.0
& \cellcolor{VLANoResult} -- & 100.0
& \cellcolor{VLANoResult} -- & 60.0
& \cellcolor{VLANoResult} -- & 100.0
& \cellcolor{VLANoResult} -- & 100.0
& \cellcolor{VLANoResult} -- & 23.3
& 70.0
& 46.7
& 43.3
& 20.0
& 73.3
& 76.7
& 67.8
\\

Octo
& \cellcolor{VLANoResult} -- & 100.0
& \cellcolor{VLANoResult} -- & 100.0
& \cellcolor{VLANoResult} -- & 63.3
& \cellcolor{VLANoResult} -- & 100.0
& \cellcolor{VLANoResult} -- & 100.0
& \cellcolor{VLANoResult} -- & 30.0
& 73.3
& 40.0
& 46.7
& 26.7
& 80.0
& 76.7
& 69.7
\\

\bottomrule

\end{tabularx}

\end{table}

\begin{expfinding}{VLMFrost}
\textbf{\textit{Findings.}} VLMs can be fine-tuned beyond their zero-shot capabilities, matching fine-tuned large-scale pretrained VLAs on complex tasks in both simulation and the real world.
\end{expfinding}

\paragraph{Fine-tuning VLMs in Simulation.}
To enable a controlled and reproducible comparison with pretrained VLAs, we select one widely used manipulation benchmark, Libero~\citep{liu2023libero}, to study the learning efficiency of fine-tuning VLMs.
On LIBERO, Table~\ref{tab:finetune-sim} compares average success rates against VLAs with large-scale action pretraining. Unlike EASG's real-world training setup, we do not use depth information in simulation; the model receives \textbf{the same input observations} as the compared VLA baselines. We use Qwen3.5-0.8B as base model. For each task, we evaluate 16 rollouts per seed over three random seeds, resulting in 48 evaluation rollouts per task. Qwen3.5-0.8B surpass Octo and OpenVLA, with a close performance to $\pi_0$-FAST.

\begin{table}[htbp]
\centering

\begin{minipage}[c]{0.80\linewidth}
\centering

\small
\setlength{\tabcolsep}{3.5pt}
\renewcommand{\arraystretch}{0.9}

\begin{tabularx}{\linewidth}{
    @{}
    >{\raggedright\arraybackslash}p{0.25\linewidth}
    >{\centering\arraybackslash}X
    >{\centering\arraybackslash}X
    >{\centering\arraybackslash}X
    >{\centering\arraybackslash}X
    >{\centering\arraybackslash}X
    @{}
}
\toprule

\textbf{Method}
& \textbf{Spatial}
& \textbf{Object}
& \textbf{Goal}
& \textbf{10}
& \textbf{Average}
\\

\midrule

\textbf{Ours(0.8B)}
& 92.5$\pm$2.08
& 93.8$\pm$1.91
& 83.1$\pm$2.96
& 63.7$\pm$3.80
& 83.3$\pm$1.40
\\

$\pi_{0}$-FAST
& 96.4
& 96.8
& 88.6
& 60.2
& \textbf{85.5}
\\

Octo
& 78.9
& 85.7
& 84.6
& 51.1
& 75.1
\\

OpenVLA
& 84.7
& 88.4
& 79.2
& 53.7
& 76.5
\\

\bottomrule

\end{tabularx}

\end{minipage}

\caption{
\textbf{Fine-tuning VLM against pretrained VLAs in simulation.}
}
\label{tab:finetune-sim}

\end{table}

\subsection{Cross-Embodiment Transfer}
\label{sec:exp-cross-embodiment}

\begin{expquestion}{VLMBlue}{
   \textit{Q3. Can VLM manipulation capabilities transfer across embodiments?}
}
\end{expquestion}

\noindent
\begin{minipage}[c]{0.67\linewidth}
Our action-space grounding represents robot actions in the end-effector pose space rather than embodiment-specific joint coordinates. This embodiment-agnostic action-space grounding naturally raises the question of whether manipulation capabilities of a fine-tuned VLM can transfer across robot embodiments.
We train the model using data collected by UR3 following the Sec.~\ref{sec:textual_action_learning}, with the exception that external-camera observations are not used in this experiment. We preserve scene configurations from Sec.~\ref{sec:exp-post-training} and replace the 6-DoF UR3 arm with a 7-DoF Flexiv robot, except external camera. We evaluate cross-embodiment transfer on Task Group 2 As shown in Table~\ref{tab:cross-embodiment}, the fine-tuned policies remain effective under the new embodiment, achieving success rates comparable to those on the original UR3 platform. For detailed cross-embodiment experimental settings, see Appendix~\ref{app:cross-embodied}.
\end{minipage}
\hfill
\begin{minipage}[c]{0.3\linewidth}
\centering

\small
\setlength{\tabcolsep}{10pt}
\renewcommand{\arraystretch}{0.95}

\begin{tabular}{lcc}
\toprule
\textbf{Task} & UR3 & Flexiv \\
\midrule
 \textbf{ 2.1} & 76.7 & 73.3 \\
 \textbf{ 2.2} & 60.0 & 56.7 \\
 \textbf{ 2.3.1} & 53.3 & 50.0 \\
 \textbf{ 2.3.2} & 36.7 & 40.0 \\
 \textbf{ 2.4.1} & 76.7 & 73.3 \\
 \textbf{ 2.4.2} & 80.0 & 80.0 \\
\midrule
\textbf{Avg.} & \textbf{63.9} & \textbf{62.2} \\
\bottomrule
\end{tabular}
\captionof{table}{
\textbf{Cross-embodiment transfer.}
}
\label{tab:cross-embodiment}
\end{minipage}

\begin{expfinding}{VLMBlue}
\textbf{\textit{Findings.}} VLM-native manipulation capabilities can transfer across embodiments with EASG.
\end{expfinding}

\section{Capability Boundaries of VLM-Native Manipulation}
\label{sec:vlm-capability-boundary}

\subsection{Inference Settings for VLM-Native Manipulation}

\begin{expquestion}{VLMCobalt}{
     \textit{\textcolor{VLMPaper}{
        Q4. How do inference configurations extend the capability frontier of VLM-native manipulation?
    }}
}
\end{expquestion}

We study two commonly available mechanisms: \textbf{memory} for interaction history across successive action predictions and \textbf{thinking} for additional reasoning. We evaluate their effects in a controlled $2\times2$
factorial study on the same real-world Tasks~1.1--1.3 with Qwen-3.8-27B. This comparison allows us
to distinguish how much performance gains arise from temporal context, additional
reasoning and both.

\begin{table}[htbp]
  \centering

  \caption{
  \textbf{Effect of memory and thinking on VLM-native zero-shot manipulation.}
  }
  \label{tab:q5-inference-settings}

  \small
  \setlength{\tabcolsep}{2.0pt}
  \renewcommand{\arraystretch}{0.82}

  \begin{tabularx}{\linewidth}{@{}l*{14}{C}@{}}
  \toprule

  &
  \multicolumn{7}{c}{\textbf{w/ Thinking}}
  &
  \multicolumn{7}{c}{\textbf{w/o Thinking}}
  \\

  \cmidrule(lr){2-8}
  \cmidrule(lr){9-15}

  \textbf{Task}
  &
  1.1.1
  & 1.1.2
  & 1.1.3
  & 1.2.1
  & 1.2.2
  & 1.3
  & \textbf{Avg.}
  &
  1.1.1
  & 1.1.2
  & 1.1.3
  & 1.2.1
  & 1.2.2
  & 1.3
  & \textbf{Avg.}
  \\

  \midrule

  w/ Memory
  &
  93.3
  & 90.0
  & 66.7
  & 100.0
  & 93.3
  & 33.3
  & 79.4
  &
  13.3
  & 16.7
  & 0.0
  & 16.7
  & 23.3
  & 0.0
  & 11.7
  \\

  w/o Memory
  &
  90.0
  & 83.3
  & 0.0
  & 100.0
  & 93.3
  & 20.0
  & 64.4
  &
  10.0
  & 6.7
  & 0.0
  & 16.7
  & 20.0
  & 0.0
  & 8.9
  \\

  \bottomrule
  \end{tabularx}
\end{table}

\begin{expfinding}{VLMCobalt}
\textbf{Findings.} Thinking is a key factor for extending VLM-native manipulation capabilities, while memory provides complementary gains mainly for longer-horizon tasks through temporal context.
\end{expfinding}

\subsection{Zero-Shot Capability Frontier of VLMs}
\label{sec:vlm-capability-boundary-zero}
\begin{expquestion}{VLMCobalt}{
     \textit{\textcolor{VLMPaper}{
        Q5. How do VLM family and scale shape the boundary of zero-shot manipulation?
    }}
}
\end{expquestion}

\begin{figure*}[htbp]
    \centering

    \begin{minipage}[c]{0.54\textwidth}
        \small
        We systematically characterize the \textbf{zero-shot manipulation frontier}
        of pretrained VLMs across model families and scales.
        We evaluate models ranging from Qwen-2B/27B/397B~\citep{qwen3_5} to larger frontier VLMs, including GPT-6 Astra and Fable~\citep{openai2026gpt6astra, anthropic2026fable51}, 
        on Tasks~1.1--1.3 under the same \method{} grounding and without any
        robot-action training.

        Rather than considering only average success rates, we examine
        \textbf{which tasks become solvable as model capability increases}.
        The task-wise scaling curves characterize how
        \textbf{model family, scale, and task complexity}
        jointly determine the boundary of VLM-native manipulation.We observe a clear scaling trend in VLM-native manipulation: task success rates consistently improve as model size increases from 2B to 27B parameters. However, further scaling from 27B to 397B yields limited gains, while the additional improvements achieved by the most advanced proprietary VLMs suggest that scaling alone may not fully determine manipulation capability.
    \end{minipage}
    \hfill
    \begin{minipage}[c]{0.45\textwidth}
        \centering
    
        \includegraphics[
            width=\linewidth
        ]{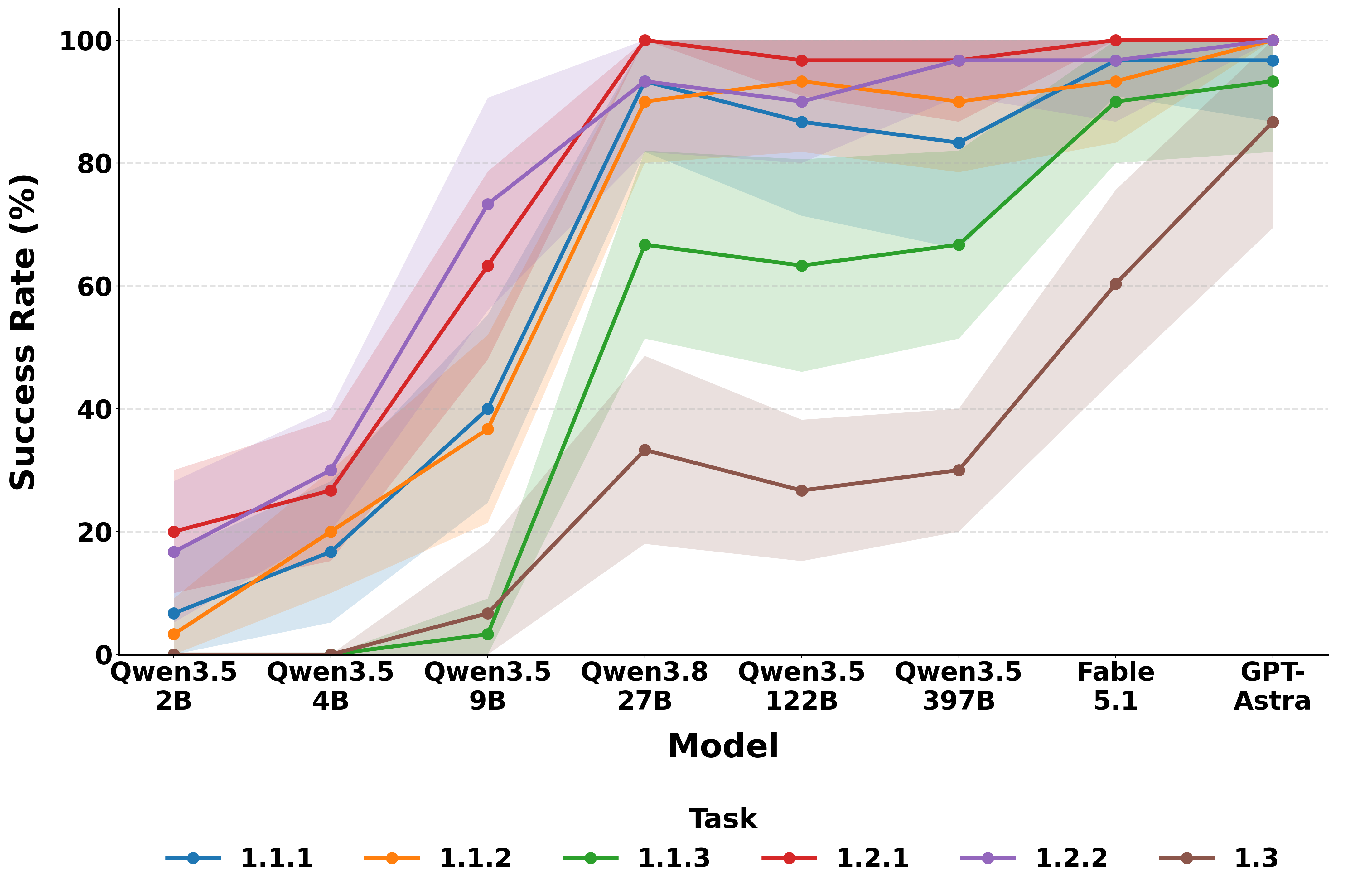}
    
        \caption{
        \textbf{Zero-shot capability frontier across VLM families and scales.}
        }
        \label{fig:zero-shot-frontier}
    
    \end{minipage}






\end{figure*}

\section{Diagnostic Ablations}
\label{sec:ablations}

\subsection{What Enables Effective Zero-Shot Action Grounding?}
\label{sec:abl-zero-shot-grounding}

\paragraph{Action-Frame and Metric Grounding.}
We ablate \textbf{action-frame grounding} by replacing or removing the TCP-centered visual frame, and \textbf{metric-depth grounding} by fixing the projected $z$-axis length and removing explicit depth information. As shown in Table~\ref{tab:q1-structure}, removing either component drops the average success rate from 79.4\% to 0\%, indicating that both are critical for effective zero-shot manipulation. Further details are provided in Appendix~\ref{app:ablation}.

\begin{table}[htbp]
  \centering
  \setlength{\belowcaptionskip}{2pt}

  \caption{
  \textbf{Effect of depth and frame grounding on real-world zero-shot
  manipulation.}
  The table reports performance on individual tasks under the same setup as experiments in Sec.~\ref{sec:exp-zero-shot-bridge}.
  }

  \label{tab:q1-structure}

  \small
  \setlength{\tabcolsep}{2.0pt}
  \renewcommand{\arraystretch}{0.82}

  \begin{tabularx}{\linewidth}{@{}l*{14}{C}@{}}
  \toprule

  &
  \multicolumn{7}{c}{\textbf{w/ Metric Depth}}
  &
  \multicolumn{7}{c}{\textbf{w/o Metric Depth}}
  \\

  \cmidrule(lr){2-8}
  \cmidrule(lr){9-15}

  \textbf{Task}
  &
  1.1.1
  & 1.1.2
  & 1.1.3
  & 1.2.1
  & 1.2.2
  & 1.3
  & \textbf{Avg.}
  &
  1.1.1
  & 1.1.2
  & 1.1.3
  & 1.2.1
  & 1.2.2
  & 1.3
  & \textbf{Avg.}
  \\

  \midrule

  w/ Frame
  &
  93.3
  & 90.0
  & 66.7
  & 100
  & 93.3
  & 33.3
  & 79.4
  &
  0.0
  & 0.0
  & 0.0
  & 0.0
  & 0.0
  & 0.0
  & 0.0
  \\

  w/o Frame
  &
  0.0
  & 0.0
  & 0.0
  & 0.0
  & 0.0
  & 0.0
  & 0.0
  &
  0.0
  & 0.0
  & 0.0
  & 0.0
  & 0.0
  & 0.0
  & 0.0
  \\

  \bottomrule

  \end{tabularx}

\end{table}


\subsection{Does GCD Improve Reliable Inference?}
\label{sec:abl-post-training}

 During post-training, VLMs must generate pose vectors through their native autoregressive decoding interface. Without additional constraints, small token-level generation errors can produce malformed or invalid action outputs. We introduce \textbf{Grammar Constrained Decoding (GCD)}, which restricts decoding to valid action formats. To isolate its effect, we fine-tune the Qwen3.5-0.8B with and without GCD and evaluate on Libero as the same setting in ~\ref{sec:exp-post-training}. As shown in Table~\ref{tab:constrained-decode}, GCD consistently improves performance across all LIBERO suites. Removing GCD reduces the average success rate from 83.28\% to 72.6\% . 

\begin{table}[H]
\centering
\small
\setlength{\tabcolsep}{4pt}
\renewcommand{\arraystretch}{0.9}

\begin{tabular}{lccccc}
\toprule
\textbf{Method}
& \textbf{Spatial}
& \textbf{Object}
& \textbf{Goal}
& \textbf{10}
& \textbf{Average}
\\
\midrule

qwen3.5-0.8B
& 92.50$\pm$2.08
& 93.75$\pm$1.91
& 83.13$\pm$2.96
& 63.75$\pm$3.80
& 83.28$\pm$1.40
\\

qwen3.5-0.8B (w.o. GCD)
& 81.25$\pm$3.09
& 78.75$\pm$3.23
& 72.50$\pm$3.53
& 58.13$\pm$3.90
& 72.66$\pm$1.73
\\

\bottomrule
\end{tabular}

\caption{
\textbf{Effect of grammar constrained decoding on simulation performance.}
}
\label{tab:constrained-decode}
\end{table}

\section{Discussions and Future Works}


We designed comprehensive experiments to validate the potential of EASG, but there are still a lot of open questions in this new native action interface waiting to be discussed. In the future, we aim to characterize zero-shot capabilities and the reasoning budgets across more open-source and proprietary VLM families and parameter scales under the same grounded action interface. It's also worth trying to extend the native textual action interface to bimanual and mobile manipulation, and verify whether the explicit action coordination is still functional as action dimensionality increases. Another direction is to investigate how scaling pretraining with the native textual action interface improves visual-spatial reasoning and zero-shot manipulation capability without adaptation using task-specific demonstrations. Hopefully, these efforts can help to further unlock the potential of VLMs as general-purpose native action models through a unified, grounded action interface.

\clearpage
\bibliography{nvlm}

\clearpage
\appendix

\renewcommand\thefigure{A\arabic{figure}}
\renewcommand\thetable{A\arabic{table}}
\renewcommand\theequation{A\arabic{equation}}
\setcounter{equation}{0}
\setcounter{table}{0}
\setcounter{figure}{0}
\startcontents[appendix]

\providecommand{\coloredappendixletter}[1]{\textcolor{VLMCobalt}{#1}}
\titlecontents{section}
  [0em]
  {\addvspace{1.2em}\large\bfseries\color{VLMCobalt}}
  {\coloredappendixletter{\thecontentslabel}\hspace{0.5em}}
  {}
  {\color{VLMCobalt}\titlerule*[0.6pc]{.}\contentspage}
\titlecontents{subsection}
  [5em]
  {\small\color{VLMBlue}}
  {\contentslabel{3.2em}}
  {}
  {\color{VLMCobalt}\titlerule*[0.55pc]{.}\contentspage}
\titlecontents{subsubsection}
  [8em]
  {\small\color{VLMBlue}}
  {\contentslabel{3.2em}}
  {}
  {\color{VLMCobalt}\titlerule*[0.55pc]{.}\contentspage}

\phantomsection
\begin{center}
  {\Large\scshape\bfseries\color{VLMCobalt} Appendix}
\end{center}
\vspace{-1.5em}
\begingroup
  \hypersetup{hidelinks}
  \setcounter{tocdepth}{3}
  \printcontents[appendix]{}{1}{}
\endgroup
\clearpage


\section{Full Related Work Analysis}
\label{app:related-work}

\subsection{Vision-Language-Action Models}
\label{app:related-work-vla}

Vision-Language-Action (VLA) models extend pretrained vision-language models (VLMs) with robot action generation and have emerged as a promising paradigm for general-purpose robotic manipulation. One family of approaches preserves the autoregressive next-token prediction interface of VLMs by \textbf{representing robot actions as discrete tokens}. RT-2 and OpenVLA discretize each scalar action dimension into 256 bins and map the resulting bin indices to tokens in the language-model vocabulary, such that a continuous robot action is generated autoregressively as a sequence of action tokens~\citep{rt22023,openvla2025}. SpatialVLA similarly predicts discrete spatial action tokens, but constructs adaptive action grids according to the empirical action distribution to better encode translation and rotation in 3D space~\citep{spatialvla2025}. FAST instead operates on action chunks: it transforms normalized action sequences into the frequency domain using a discrete cosine transform (DCT), quantizes and flattens the resulting coefficients, and further compresses them with byte-pair encoding (BPE) into compact action-token sequences~\citep{fast2025}.

Another family directly augments the VLM with \textbf{dedicated continuous-action modules}. $\pi_0$ attaches an action expert that generates continuous action chunks through conditional flow matching~\citep{pi02024, lipman2023flowmatching}, while GR00T N1 couples its vision-language module with a diffusion-transformer action module~\citep{grootn12025}. Works including OpenVLA-OFT instead replaces discrete token prediction with an MLP action head that maps VLM hidden states directly to continuous actions, using parallel decoding, action chunking~\citep{zhao2023act, diffusionpolicy2023}, and an $\ell_1$ regression objective~\citep{openvlaoft2025}. Together with these architectural advances, VLA pretraining has scaled to increasingly large and diverse robot datasets~\citep{openx2024,droid2024,robomind2024,agibotworld2025}. This paradigm integrates semantic perception, language understanding, and action prediction within a single trainable model, enabling manipulation capabilities to improve with increasing robot data and supporting multi-task manipulation, language-conditioned control, and generalization across objects and environments.

Despite this progress, the prevailing action-pretraining paradigm leaves several fundamental grounding problems to be learned implicitly from robot trajectories. First, the model must infer the \textbf{semantics of the action coordinate frame}. Cartesian translations and rotations are commonly supervised in robot-centric frames such as the robot base frame, while the relationship between this action frame and the visual observation is rarely made explicit. Consequently, the model must learn simultaneously \emph{what motion should be executed} and \emph{how that motion is represented in the coordinate system of a particular robot}. Second, the model must \textbf{recover sufficiently accurate depth information} from RGB observations for precise distance control. 

        
        


Finally, action generation requires learning \textbf{kinematic, visual-geometric, and inverse-dynamics mappings}, including mappings from joint configurations to end-effector poses and from RGB observations to executable poses. These mappings are themselves nontrivial even when treated as isolated problems, yet conventional VLA training must learn them jointly and implicitly together with task-level manipulation behavior.

As a result, substantial amounts of robot data are spent not only on learning manipulation behaviors, but also on recovering geometric relationships among observations, robot embodiments, and action spaces, leading to substantial data and training costs. Moreover, the distribution mismatch between multimodal pretraining and robot action data~\citep{driess2025knowledge,hancock2026actions} may further degrade pretrained reasoning capabilities.

\subsection{Code As Policies and Robot-Use Agents}
\label{app:related-work-code-as-policy}

There exists a group of prior works that directly exploit the native reasoning capabilities of pretrained foundation models for generalist manipulation without large-scale action pretraining. These methods generally follow a \textit{code-as-policy} paradigm, where foundation models construct explicit \textit{intermediate representations} that are subsequently converted into executable robot actions through external perception, planning, or control modules.

\paragraph{Code-as-Policy Methods.}
Code as Policies~\citep{codeaspolicies2022} formulates robot policies as executable programs generated by code-pretrained language models. The generated programs process perception outputs and compose predefined control APIs, while hierarchical code generation recursively synthesizes auxiliary functions required by more complex instructions. VoxPoser~\citep{voxposer2023} replaces explicit policy programs with composable 3D affordance and constraint value maps. Language models specify spatial objectives and constraints, which are grounded into 3D observations with visual models and subsequently optimized by a model-based motion planner to produce dense 6-DoF end-effector trajectories. CoPa~\citep{copa2024} instead grounds task-relevant object parts, identifies their spatial constraints with foundation models, and derives a sequence of task-aware 6-DoF end-effector poses that can be integrated with standard robot planning algorithms. ReKep~\citep{rekep2025} represents manipulation objectives as \textit{relational keypoint constraints}: Python functions map 3D keypoint configurations to scalar costs, and a hierarchical constrained optimizer solves these costs to generate closed-loop SE(3) end-effector trajectories. Its keypoint proposal module uses DINOv2 with
registers~\citep{dinov22024,registers2024}, while grasping
stages use AnyGrasp~\citep{anygrasp2023} to obtain grasp poses.
For collision checking, grasped rigid objects are tracked
with Cutie~\citep{cutie2024} and excluded from the scene
distance field. PIVOT~\citep{nasiriany2024pivot} enables zero-shot control
through VLM-based selection and iterative refinement of
visually annotated action proposals.

A complementary family fine-tunes VLMs to predict
\textit{image-space intermediate representations} for downstream
robot execution.
RoboPoint~\citep{yuan2025robopoint} uses a
\textit{point-based interface}, predicting spatial affordance
points that are lifted into 3D targets using depth and executed
through motion planning.
HAMSTER~\citep{li2025hamster} uses a
\textit{trajectory-based interface}, predicting coarse 2D
end-effector paths and gripper commands that guide a separately
trained 3D-aware control policy.

Across these constraint-based, point-based, and trajectory-based
interfaces, foundation models produce intermediate task
specifications, while downstream optimizers, planners, or learned
policies generate continuous robot motion. In the mean time, EASG explicitly grounds an executable continuous action space,
enabling the pretrained VLM itself to generate task-space actions
zero-shot and subsequently improve through robot-data fine-tuning
using the same action interface.

\begin{figure}[htbp]
\centering

\includegraphics[width=\textwidth]{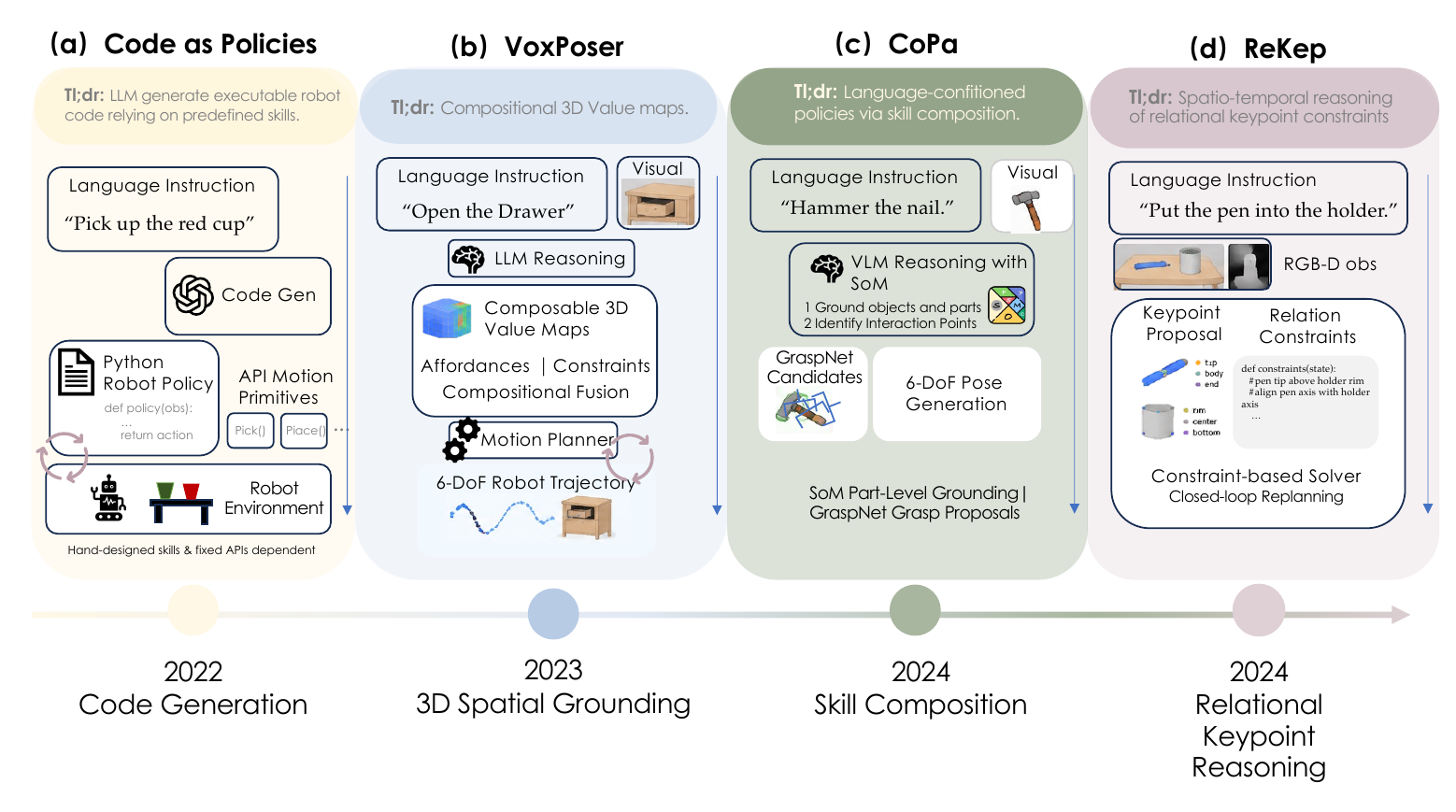}

\vspace{-0.4em}
\caption{
\textbf{Comparison with the most related methods.}
We contrast \method{} with representative prior methods to highlight the core differences in action representation, execution interface, action pretraining, zero-shot manipulation, and policy fine-tuning.
}
\label{fig:robot-use-agent-comparison}

\end{figure}

\paragraph{Robot-Use Agents.}
Recent progress in agentic foundation models further generalizes this paradigm toward systems that we broadly refer to as \textit{Robot-Use Agents}. Instead of relying on a single fixed intermediate representation, these systems expose perception, planning, execution, and robot-control modules as callable tools that can be orchestrated by a general-purpose agent. CaP-X~\citep{capx2026}, for example, provides coding agents with perception and control primitives and systematically studies different levels of abstraction, interaction, and visual grounding. Its agentic variants additionally exploit multi-turn execution feedback, visual differencing, and automatic skill synthesis, while CaP-RL post-trains the coding model with verifiable task rewards to generate more effective robot programs. VIA~\citep{via2026} takes a different approach by presenting an RGB-D-reconstructed 3D visual interface to an off-the-shelf coding agent. The agent observes screenshots, manipulates a virtual target gripper through a small set of callable tools, and delegates execution of the resulting waypoints to a low-level robot controller.

Recent systems further explore autonomous policy improvement through execution feedback. ENPIRE~\citep{enpire2026} treats a coding agent as an autonomous policy developer that can collect rollouts, diagnose failures, modify policy or training code, and invoke different policy-learning strategies through an automated robot experimentation infrastructure. ASPIRE~\citep{aspire2026} instead iteratively diagnoses failures from multimodal execution traces, synthesizes and validates program repairs, and accumulates successful repairs into an evolving reusable skill library. Importantly, these approaches introduce meaningful forms of \textit{system-level} improvement, but the learned object remains the generated program, skill scaffold, coding agent, or a separate downstream policy rather than the pretrained foundation model acting directly as a continuous-action predictor.

\paragraph{Limitations of Modular Action Interfaces.}
Such systems demonstrate that pretrained foundation models can accomplish a subset of manipulation tasks zero-shot without conventional large-scale robot-action pretraining. However, their capabilities are mediated by the surrounding tool pipeline. The foundation model does not \textbf{directly generate executable continuous robot actions}; instead, its outputs must be interpreted through \textbf{manually or automatically constructed programs}, geometric representations, motion planners, trajectory optimizers, or semantic control tools. Consequently, performance can depend strongly on the abstractions and capabilities exposed by the external scaffold.

This modular design can additionally introduce considerable \textit{\textbf{system complexity}}. Execution may require multiple perception models, geometric processing stages, planning or optimization modules, code execution environments, and robot-specific control interfaces. These dependencies can increase planning latency and engineering overhead, particularly for closed-loop manipulation requiring frequent visual feedback and action updates.

More importantly, when native zero-shot reasoning becomes insufficient, these modular pipelines lack a unified mechanism for \textbf{\textit{fine-tuning the foundation models in continuous action spaces}}. Classical code-as-policy methods largely retain fixed foundation models and improve execution through their external representations and planners~\citep{codeaspolicies2022,voxposer2023,copa2024,rekep2025}. More recent agentic systems can improve generated programs, learn reusable skills, post-train coding agents, or train separate downstream policies~\citep{capx2026,enpire2026,aspire2026}; however, these forms of adaptation do not directly turn the same pretrained foundation model used for zero-shot reasoning into a continuous robot-action policy.

We make a short summary of the above discussion and comparison in Table~\ref{tab:code-as-policy-comparison}.

\begin{table*}[htbp]
\centering
\caption{
\textbf{Comparison with representative code-as-policy and Robot-Use Agent methods.}
Prior approaches connect foundation-model reasoning to physical control through
explicit intermediate representations or external tools, whereas \method{}
directly grounds continuous robot poses into the VLM native interface.
}
\label{tab:code-as-policy-comparison}

\small
\setlength{\tabcolsep}{5pt}
\renewcommand{\arraystretch}{1.05}

\begin{tabular}{
@{}
>{\raggedright\arraybackslash}m{0.18\textwidth}
>{\raggedright\arraybackslash}m{0.17\textwidth}
>{\raggedright\arraybackslash}m{0.30\textwidth}
>{\raggedright\arraybackslash}m{0.25\textwidth}
@{}
}

\toprule
\textbf{Method} &
\textbf{Action Interface} &
\textbf{Execution Pipeline} &
\textbf{Further Improvement} \\
\midrule

Code as Policies~\citep{codeaspolicies2022}
&
Executable policy programs
&
\pipeline{
\tabitem Perception APIs\\
\tabitem Geometric operations\\
\tabitem Robot-control APIs
}
&
No foundation-model training
\\
\cmidrule(lr){1-4}

VoxPoser~\citep{voxposer2023}
&
3D affordance and constraint maps
&
\pipeline{
\tabitem RGB-D grounding\\
\tabitem 3D value-map construction\\
\tabitem Model-based motion planning
}
&
No foundation-model training
\\
\cmidrule(lr){1-4}

CoPa~\citep{copa2024}
&
Part-level spatial constraints and EEF poses
&
\pipeline{
\tabitem Object-part grounding\\
\tabitem Grasp proposal\\
\tabitem Geometric constraint solving
}
&
No foundation-model training
\\
\cmidrule(lr){1-4}

ReKep~\citep{rekep2025}
&
Relational keypoint constraints
&
\pipeline{
\tabitem Keypoint perception\\
\tabitem Constraint generation\\
\tabitem Hierarchical SE(3) optimization
}
&
No foundation-model training
\\
\cmidrule(lr){1-4}

CaP-X~\citep{capx2026}
&
Programs over perception/control primitives
&
\pipeline{
\tabitem Perception/control tools\\
\tabitem Program synthesis\\
\tabitem Execution feedback
}
&
Skills \& coding-agent RL
\\
\cmidrule(lr){1-4}

VIA~\citep{via2026}
&
Visual tool calls and target-gripper waypoints
&
\pipeline{
\tabitem RGB-D 3D interface\\
\tabitem Target-gripper control\\
\tabitem Low-level execution
}
&
No foundation-model training
\\
\cmidrule(lr){1-4}

ENPIRE~\citep{enpire2026}
&
Policy and training code
&
\pipeline{
\tabitem Robot rollouts\\
\tabitem Failure diagnosis\\
\tabitem Downstream policy training
}
&
Policy learning; no FM training
\\
\cmidrule(lr){1-4}

ASPIRE~\citep{aspire2026}
&
Policy programs and reusable skills
&
\pipeline{
\tabitem Execution tracing\\
\tabitem Program repair\\
\tabitem Skill validation
}
&
Skills \& iterative refinement
\\

\midrule

\textbf{\method{} (Ours)}
&
\textbf{Continuous pose vectors}
&
\pipeline{
\textbf{\tabitem Camera-projected grounding}\\
\textbf{\tabitem TCP-centered representation}\\
\textbf{\tabitem Direct VLM action prediction}
}
&
\pipeline{
\textbf{Continuous-action space}\\
\textbf{VLM fine-tuning}
}
\\

\bottomrule
\end{tabular}
\end{table*}

\section{\method Method Details}
\label{app:implement}

\subsection{Zero-shot Action-Space Grounding Prompts}
\label{app:prompts}

We provide the complete prompts used for zero-shot action-space grounding in this section. These prompts specify the grounded action representation, the interpretation of the projected TCP-centered coordinate frame, and the required output format for direct robot pose generation. Unless otherwise stated, the same prompt structure is used across zero-shot experiments, with only task-specific instructions and observations changed accordingly.

\begin{expquestion}{VLMSky}{
    \textit{Linguistic Action-Space Grounding}
}
Control a robotic arm using the current visual observations. The wrist camera is mounted near a parallel-jaw gripper, and the processed image marks the TCP control point and coordinate axes; these graphics are not objects to grasp.

Express all motion commands in the TCP coordinate frame attached to the gripper. Its origin is the TCP center:
\begin{itemize}
\item X (red): runs across the gap between the two fingers, along the gripper's opening and closing direction.
\item $+Z$ (blue): points forward along the fingers, from the gripper body toward the fingertips.
\item Y (green): is perpendicular to the plane formed by X and Z.
\end{itemize}

The colored arrows indicate the positive directions. The axes form a right-handed coordinate frame:
\[
+X \times +Y = +Z.
\]
Negative directions are opposite to the corresponding positive directions.

These arrows are projections of the gripper's 3D axes onto the image, not the camera's image axes. The axes are perpendicular in 3D, but their projections may appear tilted or non-perpendicular.

Output one incremental action, expressing both the translation direction and rotation axis in the TCP frame before the action is executed. Translation specifies the displacement of the TCP center; rotation specifies the orientation change about that center as a unit axis and a signed angle.

\textbf{Image order when multiple images are provided:}
\begin{enumerate}
    \item Raw wrist-camera image.
    \item Processed wrist-camera image with the TCP-center coordinate frame.
    \item External overview image, if provided: use it to understand the global scene, object identity, occlusions, and task completion, not to define motion axes.
\end{enumerate}

\textbf{Current TCP Z distance information:}

\texttt{[TCP\_Z\_DISTANCE\_INFO]}

The blue endpoint marks the visible surface point used to estimate clearance along TCP $+Z$, not a fixed axis length. The estimate may refer to a surface other than the target and does not guarantee clearance for the entire gripper. Use it together with finger-object alignment to judge grasping range.

\textbf{Current gripper state:}

\texttt{[Gripper state]}

Gripper opening ranges continuously from 0 (fully closed) to 1 (fully open). The input is the measured opening; the output \texttt{gripper\_state} is the desired opening, not a change in opening.

\textbf{Task:}

\texttt{[Task]}

Return one line in the exact format below, without spaces, unit suffixes, or extra text. \texttt{task\_state} must be the integer 0 or 1; all other values must be finite and include an explicit $+$ or $-$ sign and exactly two decimal places.

\textbf{Strict output format:}

\begin{verbatim}
task_state|direction_x,direction_y,direction_z,distance_cm|
axis_x,axis_y,axis_z,angle_deg|gripper_state
\end{verbatim}

\textbf{Rules:}
\begin{itemize}
    \item Set \texttt{task\_state} to 1 only when current observations confirm completion of the entire task; otherwise use 0.
    
    \item If \texttt{task\_state} is 1, output a no-op: zero translation, zero rotation, and an unchanged gripper state.
    
    \item For nonzero translation or rotation, normalize the direction or axis to a 3D unit vector before rounding its components.
    
    \item \texttt{distance\_cm} is the nonnegative translation distance in centimeters.
    
    \item If no translation is needed, use
    \begin{verbatim}
+0.00,+0.00,+0.00,+0.00
    \end{verbatim}
    for the translation field.
    
    \item \texttt{angle\_deg} is the signed rotation angle in degrees, following the right-hand rule around the specified axis.
    
    \item If no rotation is needed, use
    \begin{verbatim}
+0.00,+0.00,+1.00,+0.00
    \end{verbatim}
    for the rotation field.
\end{itemize}

\textbf{Example:}

\begin{verbatim}
0|+0.40,-0.10,+0.91,+6.00|+0.00,+0.00,+1.00,-45.00|+1.00
\end{verbatim}

\end{expquestion}


\subsection{Implementation of Grammar-Constrained Decoding}
\label{app:constrained-decoding}

Although the native language decoder allows the VLM to express continuous
robot actions without introducing a specialized action vocabulary, unrestricted
generation may produce malformed commands or additional natural-language text.
We therefore apply grammar-constrained autoregressive decoding at inference
time. The constraint operates only on the syntactic form of the action and
does not alter the relative logits of valid tokens.

\paragraph{Action grammar.}
For the single-arm policy, each generated action has the form
\begin{center}
\small
\texttt{
0|+0.40,-0.10,+0.91,+6.00|
+0.00,+0.00,+1.00,-45.00|+1.00
}.
\end{center}
The four fields correspond to task state, translation, rotation, and gripper
state, respectively. The translation field contains a 3D unit direction and
a distance in centimeters, while the rotation field contains a 3D unit axis
and a signed angle in degrees.

We define the valid action language using the following grammar:
\begin{equation}
\begin{aligned}
\langle\mathrm{action}\rangle
&::=
\langle\mathrm{state}\rangle
\texttt{|}
\langle\mathrm{trans}\rangle
\texttt{|}
\langle\mathrm{rot}\rangle
\texttt{|}
\langle\mathrm{grip}\rangle ,
\\
\langle\mathrm{state}\rangle
&::= \texttt{0}\mid\texttt{1},
\\
\langle\mathrm{trans}\rangle
&::=
\langle n\rangle\texttt{,}
\langle n\rangle\texttt{,}
\langle n\rangle\texttt{,}
\langle n\rangle,
\\
\langle\mathrm{rot}\rangle
&::=
\langle n\rangle\texttt{,}
\langle n\rangle\texttt{,}
\langle n\rangle\texttt{,}
\langle n\rangle,
\\
\langle\mathrm{grip}\rangle
&::= \langle n\rangle,
\\
\langle n\rangle
&::=
\langle\mathrm{sign}\rangle
\langle\mathrm{int}\rangle
\texttt{.}
\langle\mathrm{digit}\rangle
\langle\mathrm{digit}\rangle,
\\
\langle\mathrm{sign}\rangle
&::= \texttt{+}\mid\texttt{-},
\qquad
\langle\mathrm{int}\rangle ::= \langle\mathrm{digit}\rangle^{+},
\\
\langle\mathrm{digit}\rangle
&::= \texttt{0}\mid\cdots\mid\texttt{9}.
\end{aligned}
\label{eq:action_grammar}
\end{equation}
Thus, every continuous quantity must contain an explicit sign and exactly two
digits after the decimal point. The grammar enforces only serialization
syntax. Semantic constraints---such as unit-norm translation directions and
rotation axes, nonnegative translation distance, valid gripper range, and
task-completion semantics---are specified by the policy prompt and handled
independently of the decoding grammar.

\paragraph{Finite-state prefix validation.}
We compile the grammar into a deterministic finite-state machine.
Let $q_k$ denote its state after the currently generated textual prefix
$y_{<k}$, and let
\[
    \delta(q,c)
\]
denote the transition produced by consuming character $c$.
An invalid transition enters a rejecting state $\bot$.
For a character string
$s=(c_1,\ldots,c_m)$, we write
\begin{equation}
    \delta^{*}(q,s)
    =
    \delta(
        \cdots
        \delta(
            \delta(q,c_1),c_2
        ),
        \ldots,c_m
    )
\end{equation}
for the state obtained after consuming the complete string.

Importantly, the constraint is applied at the \emph{token} level rather than
assuming a one-to-one correspondence between tokens and characters.
Let $\tau(v)$ denote the text emitted by vocabulary token $v$ under the
pretrained tokenizer. A token is considered valid after prefix $y_{<k}$ iff
its entire decoded string can be consumed by the grammar:
\begin{equation}
    \mathcal{V}_{G}(q_k)
    =
    \left\{
        v\in\mathcal{V}
        \;\middle|\;
        \delta^{*}(q_k,\tau(v))\neq\bot
    \right\}.
    \label{eq:allowed_tokens}
\end{equation}
This treatment naturally supports subword tokens that span multiple grammar
symbols, such as a sign followed by digits or a delimiter followed by the
beginning of the next field.

The end-of-sequence token is permitted only when the grammar is in an accepting
state, i.e., after the second decimal digit of the final gripper value.
Whitespace, natural-language tokens, malformed separators, missing signs, and
numbers with an invalid decimal format are therefore rejected before they can
be generated.

\paragraph{Logit masking.}
At autoregressive step $k$, the VLM first produces its original vocabulary
logits $z_k(v)$. We preserve these logits for grammar-valid tokens and mask all
others:
\begin{equation}
    \widetilde{z}_k(v)
    =
    \begin{cases}
        z_k(v),
        & v\in\mathcal{V}_{G}(q_k),
        \\
        -\infty,
        & \text{otherwise}.
    \end{cases}
    \label{eq:grammar_mask_appendix}
\end{equation}
The next token is then selected from
$\operatorname{softmax}(\widetilde{\mathbf z}_k)$ using the same decoding rule
as the underlying VLM. After token $y_k$ is selected, the grammar state is
updated as
\begin{equation}
    q_{k+1}
    =
    \delta^{*}
    \left(
        q_k,\tau(y_k)
    \right).
    \label{eq:grammar_state_update}
\end{equation}
Consequently, constrained decoding changes only the support of the next-token
distribution: among all syntactically valid continuations, token preferences
remain those of the original pretrained language decoder.

\paragraph{Implementation.}
We implement the grammar as an inference-time logits processor. Generation
starts from the grammar's initial state at the first assistant-output token.
At every decoding step, the processor determines the vocabulary tokens whose
decoded strings admit a valid transition from the current grammar state,
masks all remaining logits, and advances the state using the selected token.
Because the grammar contains no learned parameters and uses only the
pretrained tokenizer vocabulary, it introduces neither an additional action
head nor robot-specific action tokens.
\label{app:formulation}

\section{Zero-shot Experiments}
\label{app:zeroshot}

\paragraph{Zero-shot Experiments.}
In our setting, zero-shot execution refers to directly deploying a pretrained VLM to perform a manipulation task without any robot-action training, using the explicit action-space grounding (\methodlit) described in Sec.~\ref{sec:method} and further detailed in Appendix~\ref{app:implement}. At each control step, the VLM receives the current visual observations, the task instruction, and explicit action-space grounding that combines visual, metric, and linguistic cues. It then autoregressively outputs a robot action represented entirely in plain text. The textual action specifies task completion, incremental translation, incremental rotation, and gripper control in the grounded action frame. The predicted action is executed, after which the environment is re-observed and the process repeats in closed loop.

\subsection{Experiment Settings}
\label{app:real-zero-exp}

\paragraph{Real-World Hardware Setup.} Our real-world experiment platform consists of a Universal Robots UR3 arm with a Robotiq 2F-85 gripper, operating over a $60\times60$\,cm tabletop workspace. We use a wrist-mounted Orbbec Gemini 305 camera and a fixed external Intel RealSense D435i camera to provide visual observations. Unless otherwise stated (for example, in the experiment of cross-embodiment in Sec.~\ref{sec:exp-cross-embodiment}), all real-world experiments share the same robot, camera configuration, and workspace, both the tasks tested in the zero-shot experiments and the tasks tested in the fine-tuning experiments.

\begin{figure}[htbp]
    \centering

    \includegraphics[width=0.8\linewidth]{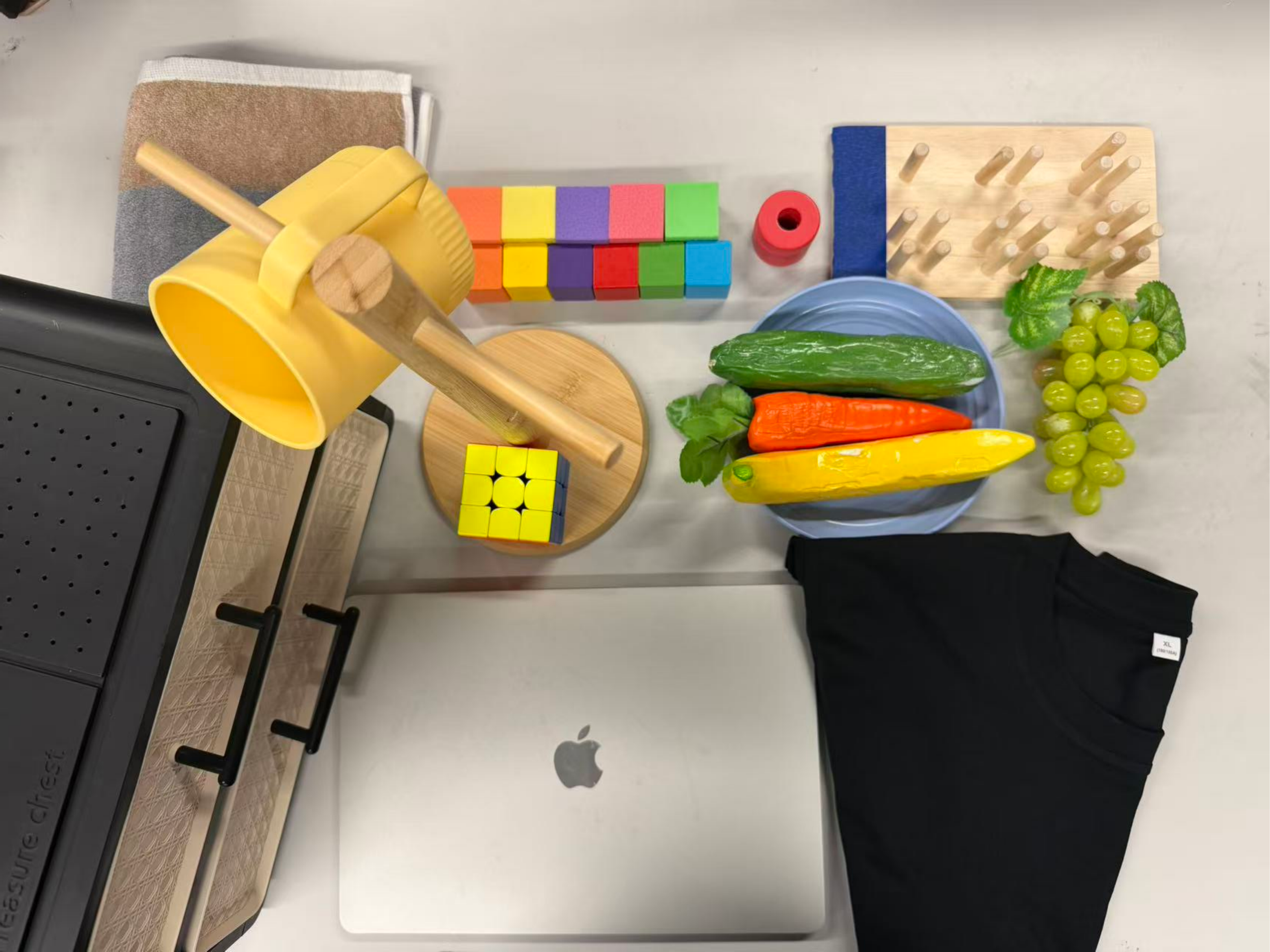}

    \vspace{-0.3em}
    \caption{\textbf{Objects used in our real-world manipulation experiments.}}\label{fig:real-world-task-objects}

    \label{fig:real-world-task-objects}
\end{figure}

\paragraph{Task Composition Analysis.}
We construct a diverse set of real-world manipulation tasks covering rigid-object manipulation, deformable-object manipulation, articulated-object interaction, precise/contact-rich manipulation, and long-horizon multi-stage tasks. The task definitions and corresponding success criteria are summarized in Table ~\ref{tab:real-tasks}.

















\begin{table}[htbp]
\vspace{-0.5em}
\centering

\caption{
\textbf{Task Group 1: Zero-shot evaluation tasks (Easy).}
}
\label{tab:real-tasks}

\vspace{-0.8em}

\footnotesize
\setlength{\tabcolsep}{3.0pt}
\renewcommand{\arraystretch}{0.85}

\begin{tabularx}{\linewidth}{
    @{}
    >{\centering\arraybackslash}p{0.17\linewidth}
    >{\centering\arraybackslash}p{0.10\linewidth}
    >{\raggedright\arraybackslash}p{0.30\linewidth}
    X
    @{}
}

\toprule

\textbf{Type}
& \textbf{Task}
& \textbf{Task description}
& \textbf{Success criterion}
\\

\midrule

\multirow{3}{*}{\raisebox{-1.5ex}{Rigid}}
& 1.1.1
& Grasp the cucumber
& The cucumber is successfully grasped and held by the robot.
\\

\cmidrule(lr){2-2}
\cmidrule(lr){3-4}

&
1.1.2
& Pick up the block
& The block is lifted off the table and held by the robot.
\\

\cmidrule(lr){2-2}
\cmidrule(lr){3-4}

&
1.1.3
& Pick up the banana and put it on the plate
& The banana is lifted from the table and placed on the plate.
\\

\cmidrule(lr){1-1}
\cmidrule(lr){2-2}
\cmidrule(lr){3-4}

\multirow{2}{*}{Articulated}
& 1.2.1
& Close the laptop
& The laptop lid reaches the closed position.
\\

\cmidrule(lr){2-2}
\cmidrule(lr){3-4}

&
1.2.2
& Push the drawer closed
& The drawer is fully closed.
\\

\cmidrule(lr){1-1}
\cmidrule(lr){2-2}
\cmidrule(lr){3-4}

Deformable
& 1.3
& Fold the towel
& The towel is folded with partial overlap between folded regions.
\\

\bottomrule

\end{tabularx}

\vspace{-0.8em}

\end{table}

\begin{figure}[htbp]
    \centering
    \includegraphics[width=0.95\linewidth]{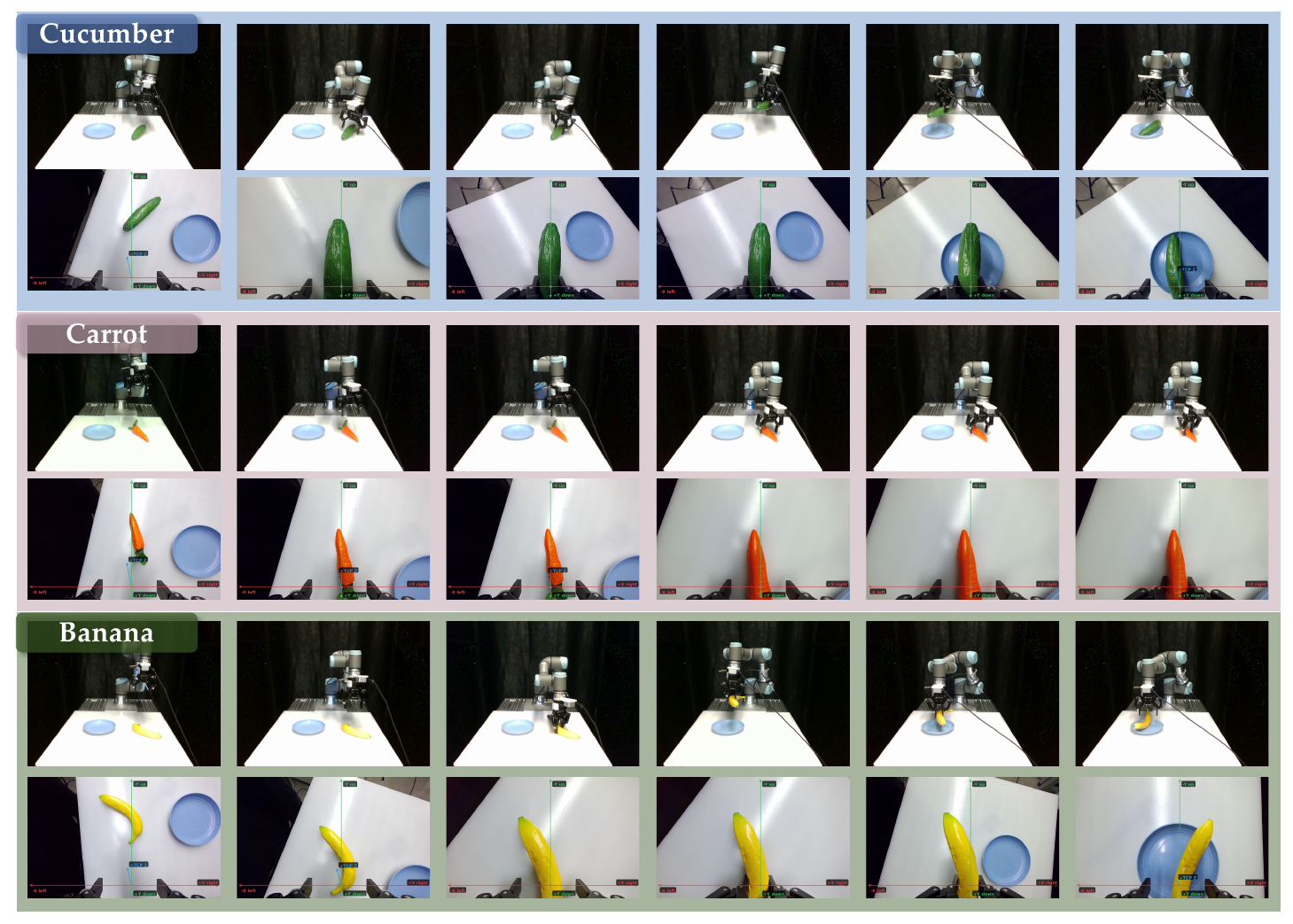}

    \vspace{-0.4em}
    \caption{
        \textbf{Demonstration of successful zero-shot rollout cases.}
    }
    \label{fig:inference-cases}
\end{figure}

\paragraph{Representative Zero-shot Inference Trajectories.}
For each task, we select one successful zero-shot trajectory according to the task-specific success metrics defined in Table~\ref{tab:real-tasks}. Figure~\ref{fig:inference-cases} visualizes these representative trajectories.

\subsection{Baseline Models}
\label{app:real-zero-baseline}

\paragraph{VLA Baselines.}
We compare \method{} against $\pi_{0.5}$, GR00T N1.6, $\pi_0$-FAST, and OpenVLA-OFT, representing recent general-purpose VLA policies with different model architectures and action-generation schemes. Each baseline is fine-tuned on task-specific demonstrations collected with the same embodiment, workspace, and task definitions used for evaluating \method{}. As presented in Table ~\ref{tab:zeroshot-vla-finetune-full}, we vary the amount of fine-tuning data from 10 to 50 demonstrations per task to measure how much robot data is required for these pretrained VLAs to match the zero-shot performance of \method{}. Detailed results across all data budgets are reported below.

\begin{table}[htbp]
\centering
\vspace{-0.5em}

\caption{
\textbf{Full comparison with different amounts of robot data.}
We report mean success rates $\pm$ standard deviations for VLA baselines
fine-tuned with 10/20/30/40/50 demonstrations per task.
Ours uses no task-specific demonstrations.
}
\label{tab:zeroshot-vla-finetune-full}

\vspace{-0.8em}

\scriptsize
\renewcommand{\arraystretch}{0.88}

\resizebox{\linewidth}{!}{
\begin{tabular}{
    @{}
    >{\raggedright\arraybackslash}p{1.6cm}
    c
    *{7}{c}
}
\toprule

\textbf{Method}
& \textbf{Demos.}
& \textbf{1.1.1}
& \textbf{1.1.2}
& \textbf{1.1.3}
& \textbf{1.2.1}
& \textbf{1.2.2}
& \textbf{1.3}
& \textbf{Avg.}
\\

\midrule


\rowcolor{VLMFrost}
\cellcolor{white}\textbf{Ours}
& \textit{Zero}
& $93.3\pm5.8$
& $90.0\pm10.0$
& $66.7\pm15.3$
& $100.0\pm0.0$
& $93.3\pm5.8$
& $33.3\pm15.3$
& $79.4\pm2.5$
\\

\midrule


\rowcolor{VLMCream!50}
\cellcolor{white}
& 10
& $70.0\pm10.0$
& $66.7\pm15.3$
& $30.0\pm10.0$
& $50.0\pm10.0$
& $46.7\pm15.3$
& $0.0\pm0.0$
& $43.9\pm2.5$
\\

\rowcolor{VLMCream}
\cellcolor{white}
& 20
& $86.7\pm15.3$
& $83.3\pm15.3$
& $56.7\pm15.3$
& $66.7\pm15.3$
& $56.7\pm15.3$
& $13.3\pm15.3$
& $60.6\pm2.5$
\\

\rowcolor{VLMFrost}
\cellcolor{white}$\pi_{0.5}$
& 30
& $100.0\pm0.0$
& $100.0\pm0.0$
& $80.0\pm10.0$
& $80.0\pm10.0$
& $66.7\pm15.3$
& $30.0\pm10.0$
& $76.1\pm2.5$
\\

\rowcolor{VLMSky}
\cellcolor{white}
& 40
& $100.0\pm0.0$
& $100.0\pm0.0$
& $93.3\pm5.8$
& $93.3\pm5.8$
& $90.0\pm10.0$
& $66.7\pm15.3$
& $90.6\pm2.5$
\\

\rowcolor{VLMSky!120}
\cellcolor{white}
& 50
& $100.0\pm0.0$
& $100.0\pm0.0$
& $100.0\pm0.0$
& $100.0\pm0.0$
& $100.0\pm0.0$
& $80.0\pm10.0$
& $96.7\pm1.7$
\\

\midrule


\rowcolor{VLMCream!50}
\cellcolor{white}
& 10
& $53.3\pm15.3$
& $53.3\pm15.3$
& $0.0\pm0.0$
& $46.7\pm15.3$
& $30.0\pm10.0$
& $0.0\pm0.0$
& $30.6\pm2.5$
\\

\rowcolor{VLMCream}
\cellcolor{white}
& 20
& $73.3\pm15.3$
& $70.0\pm10.0$
& $23.3\pm15.3$
& $50.0\pm10.0$
& $43.3\pm15.3$
& $3.3\pm5.8$
& $43.9\pm2.5$
\\

\rowcolor{VLMFrost}
\cellcolor{white}$\pi_{0}$-FAST
& 30
& $90.0\pm10.0$
& $86.7\pm15.3$
& $46.7\pm15.3$
& $53.3\pm15.3$
& $53.3\pm15.3$
& $10.0\pm10.0$
& $56.7\pm2.9$
\\

\rowcolor{VLMSky}
\cellcolor{white}
& 40
& $96.7\pm5.8$
& $96.7\pm5.8$
& $66.7\pm15.3$
& $86.7\pm15.3$
& $90.0\pm10.0$
& $30.0\pm10.0$
& $77.8\pm2.5$
\\

\rowcolor{VLMSky!120}
\cellcolor{white}
& 50
& $100.0\pm0.0$
& $100.0\pm0.0$
& $73.3\pm15.3$
& $100.0\pm0.0$
& $100.0\pm0.0$
& $40.0\pm10.0$
& $85.6\pm2.5$
\\

\midrule


\rowcolor{VLMCream!50}
\cellcolor{white}
& 10
& $66.7\pm15.3$
& $60.0\pm10.0$
& $10.0\pm10.0$
& $50.0\pm10.0$
& $46.7\pm15.3$
& $0.0\pm0.0$
& $38.9\pm2.5$
\\

\rowcolor{VLMCream}
\cellcolor{white}
& 20
& $86.7\pm15.3$
& $83.3\pm15.3$
& $36.7\pm15.3$
& $53.3\pm15.3$
& $50.0\pm10.0$
& $13.3\pm15.3$
& $53.9\pm2.5$
\\

\rowcolor{VLMFrost}
\cellcolor{white}GR00T N1.6
& 30
& $100.0\pm0.0$
& $100.0\pm0.0$
& $70.0\pm10.0$
& $53.3\pm15.3$
& $50.0\pm10.0$
& $33.3\pm15.3$
& $67.8\pm2.5$
\\

\rowcolor{VLMSky}
\cellcolor{white}
& 40
& $100.0\pm0.0$
& $100.0\pm0.0$
& $93.3\pm5.8$
& $90.0\pm10.0$
& $90.0\pm10.0$
& $60.0\pm10.0$
& $88.9\pm2.5$
\\

\rowcolor{VLMSky!120}
\cellcolor{white}
& 50
& $100.0\pm0.0$
& $100.0\pm0.0$
& $100.0\pm0.0$
& $100.0\pm0.0$
& $100.0\pm0.0$
& $73.3\pm15.3$
& $95.6\pm2.5$
\\

\midrule


\rowcolor{VLMCream!50}
\cellcolor{white}
& 10
& $66.7\pm15.3$
& $66.7\pm15.3$
& $20.0\pm10.0$
& $53.3\pm15.3$
& $40.0\pm10.0$
& $0.0\pm0.0$
& $41.1\pm2.5$
\\

\rowcolor{VLMCream}
\cellcolor{white}
& 20
& $86.7\pm15.3$
& $83.3\pm15.3$
& $36.7\pm15.3$
& $60.0\pm10.0$
& $56.7\pm15.3$
& $6.7\pm5.8$
& $55.0\pm2.9$
\\

\rowcolor{VLMFrost}
\cellcolor{white}OpenVLA-OFT
& 30
& $100.0\pm0.0$
& $93.3\pm5.8$
& $53.3\pm15.3$
& $66.7\pm15.3$
& $70.0\pm10.0$
& $13.3\pm15.3$
& $66.1\pm2.5$
\\

\rowcolor{VLMSky}
\cellcolor{white}
& 40
& $100.0\pm0.0$
& $100.0\pm0.0$
& $83.3\pm15.3$
& $90.0\pm10.0$
& $93.3\pm5.8$
& $30.0\pm10.0$
& $82.8\pm2.5$
\\

\rowcolor{VLMSky!120}
\cellcolor{white}
& 50
& $100.0\pm0.0$
& $100.0\pm0.0$
& $93.3\pm5.8$
& $100.0\pm0.0$
& $100.0\pm0.0$
& $40.0\pm10.0$
& $88.9\pm2.5$
\\

\bottomrule
\end{tabular}
}

\vspace{-1.0em}
\end{table}

\subsection{Randomization Settings}
\label{app:real-zero-random}

\paragraph{Object and Pose Randomization.}
To evaluate zero-shot generalization beyond the configurations used in the original tasks, we replace the original object in Tasks~1.1--1.3 with five unseen objects and evaluate each method under three randomized configurations for each object. Object positions and poses are independently varied within the $60\times60$\,cm workspace while preserving the task semantics and success criteria.

Importantly, the demonstrations used to fine-tune the VLA baselines are
themselves collected with randomized object positions and poses.
Nevertheless, as reported in
Table~\ref{tab:zeroshot-randomization}, the fine-tuned VLA baselines still
exhibit a substantial performance drop when evaluated on unseen objects and
configurations, whereas \method{} largely preserves its zero-shot performance.

\begin{figure}[htbp]
    \centering

    \includegraphics[width=0.98\linewidth]{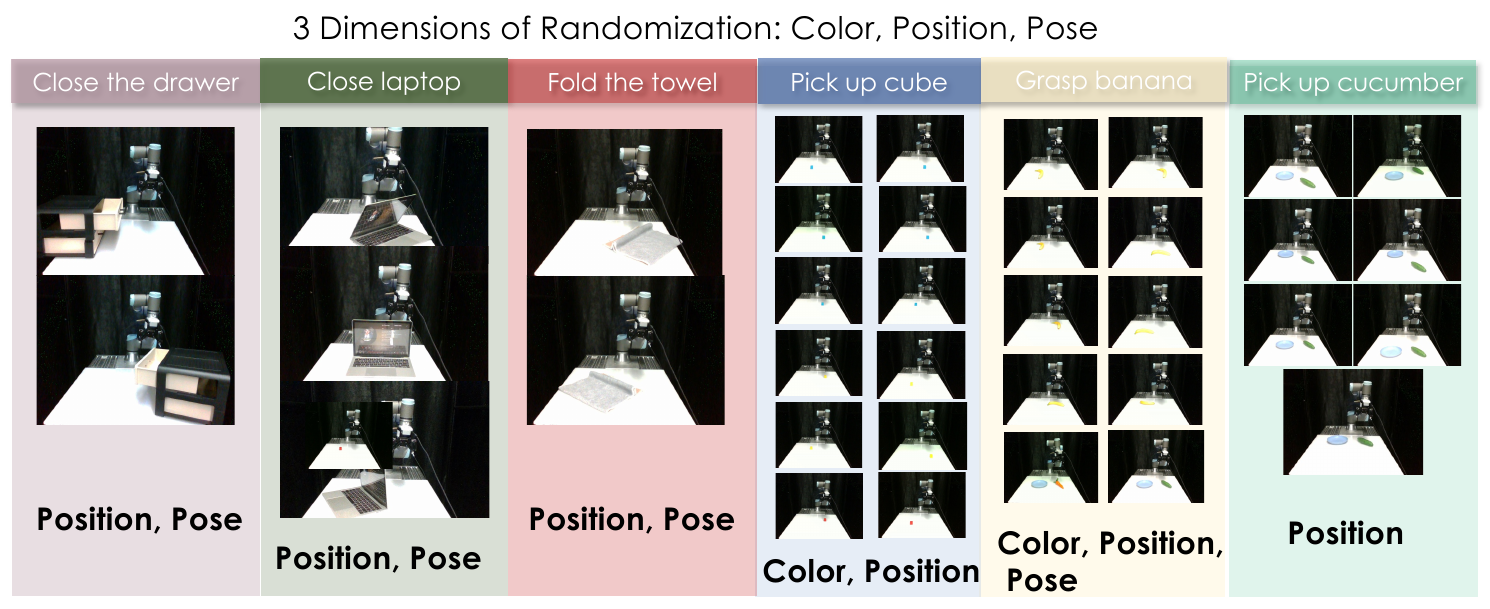}

    \vspace{-0.4em}
    \caption{
    \textbf{Randomization settings for real-world generalization experiments.}
    For each task, we vary both the object instance and its initial position
    to evaluate the robustness of \method{} beyond the original zero-shot setting.
    }
    \label{fig:randomization-settings}
\end{figure}

\section{Fine-tuning Protocols of VLMs with \method}
\label{app:finetune}

\subsection{Tasks and Metrics of Simulation Benchmarks}
\label{app:finetune-task-sim}

Simulation benchmarks and standardized evaluation platforms have become an important tool for reproducible robot-policy evaluation~\citep{robomimic2022,li24simpler,chen2025robotwin, chen2026robodojo}. We evaluate VLMs post-training with \method on \textbf{LIBERO}~\citep{liu2023libero}, which is one of the most commonly used evaluation benchmarks for VLA.
Their task compositions cover different aspects of visuomotor manipulation
and therefore provide a diverse testbed for evaluating continuous-action
fine-tuning.

\paragraph{LIBERO.}
LIBERO organizes manipulation tasks into four complementary suites:
\textbf{Spatial}, \textbf{Object}, \textbf{Goal}, and
\textbf{Long} (LIBERO-10).

\paragraph{Task Success Metrics.} For LIBERO we report \textbf{episode-level task success rate} as the primary evaluation metric. In LIBERO, an episode is considered successful once the language-conditioned task goal specified by the environment is satisfied, and success rate is computed as the fraction of successful evaluation rollouts.

\subsection{Tasks and Metrics of Real-world Scenes}
\label{app:finetune-task-real}

\paragraph{Task Composition Analysis.}
All fine-tuning tasks are intentionally designed to be more challenging than their zero-shot counterparts, targeting settings that are difficult to complete zero-shot or would otherwise require a very long time to complete in a zero-shot way. The full task compositions and success criteria are summarized in Table~\ref{tab:real-tasks-finetune}.

\begin{table}[htbp]
\vspace{-0.5em}
\centering

\footnotesize
\setlength{\tabcolsep}{4.0pt}
\renewcommand{\arraystretch}{0.82}

\begin{tabularx}{\linewidth}{
    @{}
    >{\centering\arraybackslash}p{0.13\linewidth}
    >{\centering\arraybackslash}p{0.08\linewidth}
    >{\raggedright\arraybackslash}p{0.36\linewidth}
    X
    @{}
}
\toprule

\textbf{Type}
& \textbf{Task}
& \textbf{Task description}
& \textbf{Success criterion}
\\

\midrule

Deformable
& 2.1
& Fold the T-shirt in half.
& The T-shirt is folded with the two sides partially aligned.
\\

\cmidrule(lr){1-1}
\cmidrule(lr){2-2}
\cmidrule(lr){3-4}

Articulated
& 2.2
& Open the laptop.
& The laptop lid is opened beyond 60° and remains stable.
\\

\cmidrule(lr){1-1}
\cmidrule(lr){2-2}
\cmidrule(lr){3-4}

\multirow{2}{*}{Precise}
& 2.3.1
& Pick up the ring and put it on one of the pegs.
& The ring is placed onto a peg.
\\

\cmidrule(lr){2-2}
\cmidrule(lr){3-4}

&
2.3.2
& Hang the cup on the peg.
& The cup is stably hung on the peg.
\\

\cmidrule(lr){1-1}
\cmidrule(lr){2-2}
\cmidrule(lr){3-4}

\multirow{2}{*}{Long-horizon}
& 2.4.1
& Open the drawer, then put the Rubik's Cube in the drawer.
& The drawer is partially opened and the cube is placed inside.
\\

\cmidrule(lr){2-2}
\cmidrule(lr){3-4}

&
2.4.2
& Put the grapes in the drawer, then close the drawer.
& The grapes are placed inside the drawer, and the drawer is closed within 2 cm of full closure.
\\

\bottomrule
\end{tabularx}

\vspace{-0.5em}
\caption{
\textbf{Task Group 2: Capability extension tasks (Hard).}
See Appendix~\ref{app:real-zero-exp} for the full object group we included.
}
\label{tab:real-tasks-finetune}
\vspace{-0.7em}

\end{table}

\paragraph{Real-world Success Metrics.}
The task-specific success criteria for all real-world fine-tuning tasks are summarized in Table~\ref{tab:real-tasks-finetune}. For each task and method, we conduct three independent evaluation rounds with 10 trials per round, resulting in 30 trials in total. We report the \textbf{mean task success rate $\pm$ standard deviation} across the three evaluation rounds. Tasks~1.1--1.3 are tasks that VLM with \method can complete zero-shot directly and are used to evaluate whether fine-tuning can further improve success rates over already available zero-shot capabilities. Tasks~2.1--2.4 evaluate whether \textit{\method{} fine-tuning} can help to solve harder tasks beyond the native zero-shot frontier. The complete fine-tuning results are reported in Table~\ref{tab:finetune-real-result-group2}.

\begin{table}[htbp]
\centering

\caption{
\textbf{Fine-tuning results on harder real-world manipulation tasks.}
Each entry reports the \textbf{mean task success rate $\pm$ standard deviation}
over three independent evaluation rounds with 10 trials per round.
}

\label{tab:finetune-real-result-group2}

\vspace{-0.5em}
\small
\setlength{\tabcolsep}{4.0pt}
\renewcommand{\arraystretch}{1.00}

\newcolumntype{F}{
    >{\columncolor{VLMSky}\centering\arraybackslash}X
}
\newcolumntype{A}{
    >{\columncolor{VLMSky!140}\centering\arraybackslash}X
}

\begin{tabularx}{\linewidth}{
@{}
>{\raggedright\arraybackslash}p{0.16\linewidth}
*{6}{F}
A
@{}
}

\toprule

\textbf{Method}
&
\textbf{2.1}
&
\textbf{2.2}
&
\textbf{2.3.1}
&
\textbf{2.3.2}
&
\textbf{2.4.1}
&
\textbf{2.4.2}
&
\textbf{Avg.}
\\

\midrule

\textbf{Ours}
&
\makecell{80.0\\$\pm$10.0}
&
\makecell{63.3\\$\pm$11.5}
&
\makecell{56.7\\$\pm$15.3}
&
\makecell{43.3\\$\pm$15.3}
&
\makecell{76.7\\$\pm$5.8}
&
\makecell{83.3\\$\pm$11.5}
&
\makecell{\textbf{67.2}\\$\pm$\textbf{5.1}}
\\

$\pi_{0}$-FAST
&
\makecell{83.3\\$\pm$5.8}
&
\makecell{70.0\\$\pm$10.0}
&
\makecell{53.3\\$\pm$15.3}
&
\makecell{50.0\\$\pm$10.0}
&
\makecell{80.0\\$\pm$10.0}
&
\makecell{83.3\\$\pm$5.8}
&
\makecell{70.0\\$\pm$3.3}
\\

OpenVLA
&
\makecell{70.0\\$\pm$10.0}
&
\makecell{46.7\\$\pm$15.3}
&
\makecell{43.3\\$\pm$11.5}
&
\makecell{20.0\\$\pm$10.0}
&
\makecell{73.3\\$\pm$11.5}
&
\makecell{76.7\\$\pm$5.8}
&
\makecell{55.0\\$\pm$3.3}
\\

Octo
&
\makecell{73.3\\$\pm$5.8}
&
\makecell{40.0\\$\pm$10.0}
&
\makecell{46.7\\$\pm$11.5}
&
\makecell{26.7\\$\pm$15.3}
&
\makecell{80.0\\$\pm$10.0}
&
\makecell{76.7\\$\pm$15.3}
&
\makecell{57.2\\$\pm$4.2}
\\

\bottomrule
\end{tabularx}

\end{table}

\subsection{Real-world Training Protocol}
\label{app:finetune-real}

\paragraph{Data Collection Setups.}
Real-world demonstrations are collected on a \textbf{UR3 robot} equipped with a \textbf{Robotiq 2F-85 gripper}. We use a \textbf{Haptic Touch} device for teleoperation, an \textbf{Intel RealSense D435i} as the external scene camera, and an \textbf{Orbbec Gemini 305} mounted on the robot wrist for close-range visual observations. During teleoperation, we synchronously record RGB observations, robot proprioceptive states, end-effector poses, and gripper commands at 20~Hz. The operator controls the end-effector through teleoperation mapping. Recorded trajectories are subsequently filtered and converted into the camera-projected, TCP-centered action representation used by \method{} for fine-tuning. We collect 30 demonstrations in the real-world fine-tuning experiments \textbf{(Task Group 2)} per task unless otherwise specified.







\begin{figure*}[htbp]
    \centering

    \includegraphics[width=\textwidth]{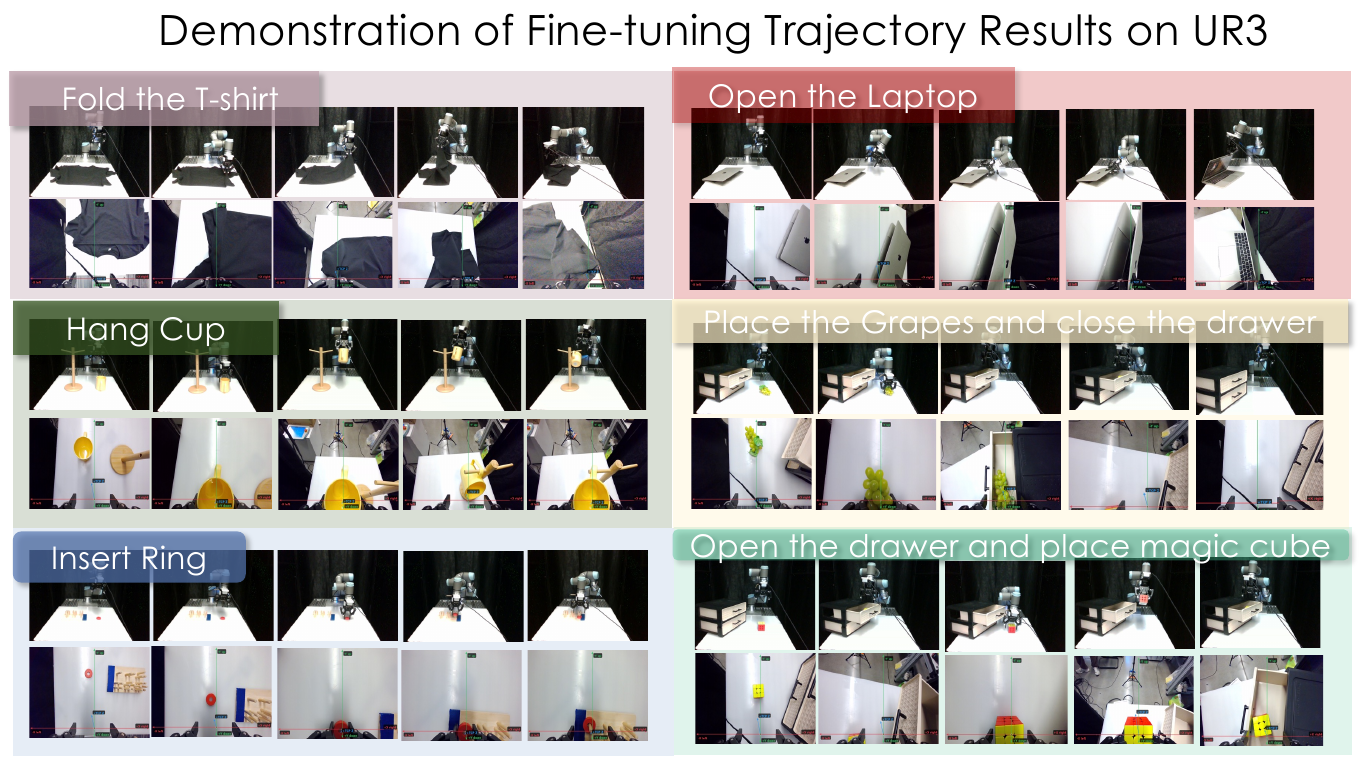}

    \caption{
    \textbf{Representative inference trajectories after \method{} fine-tuning.}
    We show one successful rollout for each Task Group~2 task using the fine-tuned Qwen-3.8-27B. These tasks are beyond the model's native zero-shot capability and are solved after continuous-action fine-tuning with \method{}.
    }
    \label{fig:finetune-real-trajectories}
\end{figure*}

\paragraph{Representative Fine-tuning Inference Trajectories.}
To qualitatively illustrate the behaviors acquired through post-training, we visualize representative successful inference trajectories of our fine-tuned Qwen3.8-27B by Lora on each \textbf{Task Group 2} task in Figure~\ref{fig:finetune-real-trajectories}. These tasks are designed beyond the model's native zero-shot capability frontier and require more precise spatial reasoning, object orientation control, or longer-horizon action sequence generation. The trajectories show that, after fine-tuning with \method{}, the VLM can reliably complete manipulation tasks beyond their zero-shot capabilities.

\section{Cross-Embodiment}
\label{app:cross-embodied}

\subsection{Experiment Design}
\label{app:cross-embodied-exp}

\paragraph{Robot Platforms.} In addition to the UR3 platform described in Appendix~\ref{app:finetune-real}, we evaluate \method{} on a second robot embodiment from Flexiv. Both platforms use the same Robotiq 2F-85 gripper, while differing substantially in arm morphology, kinematics, and low-level control. This setting allows us to isolate whether the action representation learned by \method{} transfers across robot embodiments rather than overfitting to the kinematics of a particular arm.

\begin{figure*}[htbp]
    \centering

    \includegraphics[width=0.75\textwidth]{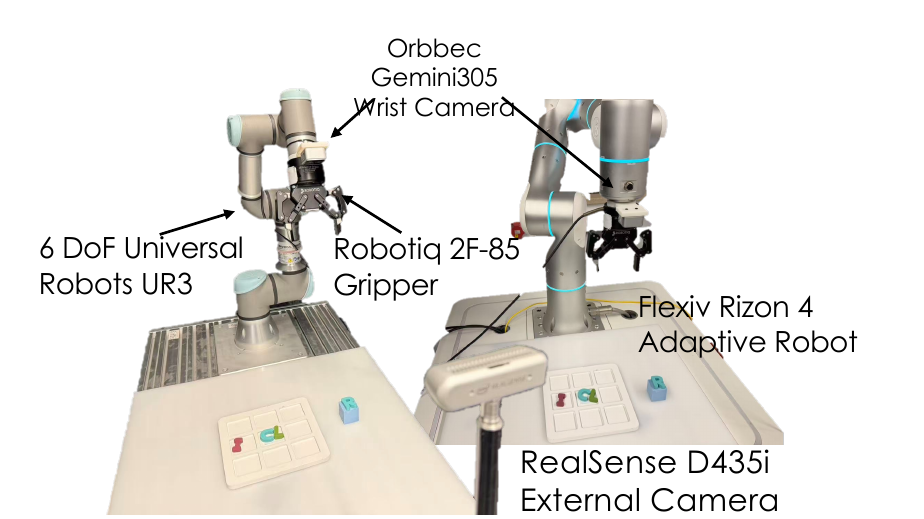}

    \caption{
    \textbf{Robot platforms used for cross-embodiment evaluation.}
    We compare the UR3 platform used for fine-tuning with a Flexiv robot arm equipped with the same Robotiq 2F-85 gripper. The two robots differ in embodiment and kinematics, while evaluation is initialized from matched end-effector poses. The external camera is only used for recording, \textbf{not for inference}.
    }
    \label{fig:cross-embodiment-rollouts}
\end{figure*}

\paragraph{Experiment Design.}
A central motivation of \method{} is that the VLM predicts \textbf{end-effector poses in a task-space representation}, rather than embodiment-specific joint commands. We therefore test whether the same fine-tuned checkpoint can be transferred directly from UR3 to Flexiv without any additional fine-tuning or embodiment-specific robot demonstrations. For each Task in Table~\ref{tab:real-tasks-finetune}, the two robots are initialized with matched end-effector poses relative to the workspace, and the same VLM checkpoint and action-grounding interface are used for inference. 

\paragraph{Cross-Embodiment Performance.}
We follow the same real-world evaluation protocol described in Appendix~\ref{app:finetune-real}, with three independent rounds of 10 trials per task. Table~\ref{tab:cross-embodiment-full} reports the \textbf{mean success rate $\pm$ standard deviation} on the original UR3 embodiment and the unseen Flexiv embodiment.

\begin{table}[htbp]
\centering

\caption{
\textbf{Cross-embodiment transfer of the same fine-tuned \method{} checkpoint.}
We report mean $\pm$ standard deviation over 3 evaluation rounds,
with 10 trials per task in each round.
No additional fine-tuning is performed for the Flexiv embodiment.
}
\label{tab:cross-embodiment-full}
\vspace{-1.0em}
\small
\setlength{\tabcolsep}{5pt}
\renewcommand{\arraystretch}{0.95}

\begin{tabularx}{\linewidth}{@{}lCCCCCCC@{}}
\toprule

\textbf{Embodiment}
& \textbf{2.1}
& \textbf{2.2}
& \textbf{2.3.1}
& \textbf{2.3.2}
& \textbf{2.4.1}
& \textbf{2.4.2}
& \textbf{Avg.}
\\

\midrule

UR3
& 76.7$\pm$11.5
& 60.0$\pm$10.0
& 53.3$\pm$11.5
& 36.7$\pm$15.3
& 76.7$\pm$5.8
& 80.0$\pm$10.0
& \textbf{63.9$\pm$5.1}
\\

Flexiv
& 73.3$\pm$11.5
& 56.7$\pm$15.3
& 50.0$\pm$10.0
& 40.0$\pm$17.3
& 73.3$\pm$5.8
& 80.0$\pm$17.3
& \textbf{62.2$\pm$3.5}
\\

\bottomrule
\end{tabularx}

\end{table}
\begin{figure*}[htbp]
    \centering

    \includegraphics[width=0.96\textwidth]{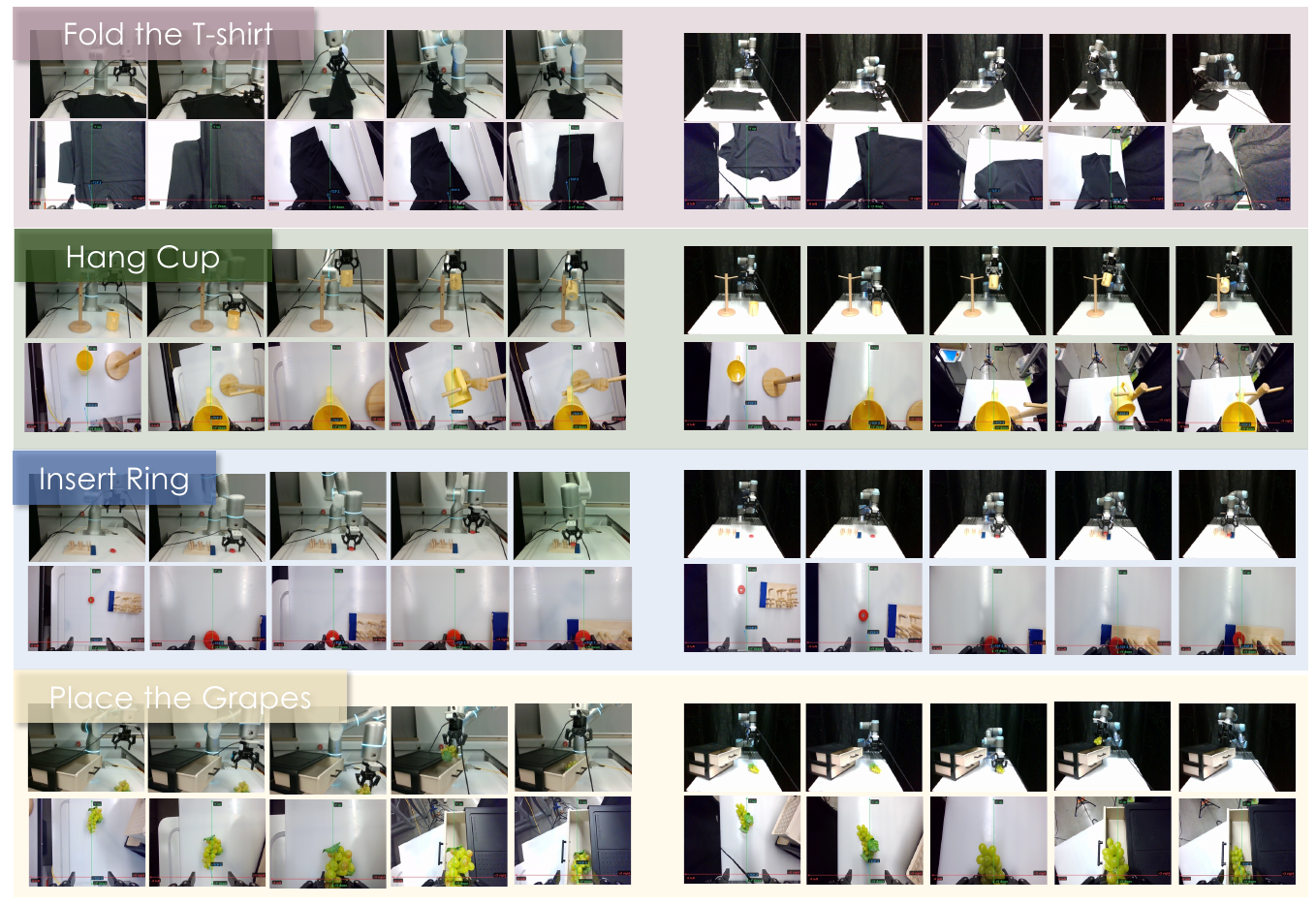}

    \caption{
    \textbf{Representative synchronized rollouts across robot embodiments.}
    We visualize the same manipulation task executed by UR3 and Flexiv using the identical fine-tuned \method{} checkpoint. Starting from matched end-effector poses, both robots receive pose predictions from the same VLM, while embodiment-specific inverse kinematics and low-level controllers execute the predicted task-space actions.
    }
    \label{fig:cross-embodiment-rollouts}
\end{figure*}
\section{Full Analysis of Ablation Study}
\label{app:ablation}

\subsection{What Enables Effective Zero-Shot Action-Space Grounding?}
\label{app:abl-zero-shot-grounding}

All zero-shot ablations in this section are conducted on \textbf{Task Group~1}
under the same robot setup, VLM backbone, prompting protocol, and inference
configuration. We modify only the corresponding grounding component to isolate
its contribution.

\paragraph{Action-Frame Grounding.}
The proposed \method{} represents actions in a \textbf{TCP-centered coordinate
frame}, such that the predicted motion is expressed relative to the controllable
end effector. To examine whether this choice is important, we replace the
TCP-centered frame with two alternative coordinate systems: a
\textbf{robot-base-centered frame} and a \textbf{wrist-camera-centered frame}, while
keeping all other components unchanged. This comparison tests whether the zero-shot capability arises specifically from aligning the action representation
with the locally controllable tool frame, rather than merely exposing an
arbitrary robot coordinate system to the VLM.

\begin{figure}[htbp]
    \centering
    \includegraphics[width=0.95\linewidth]{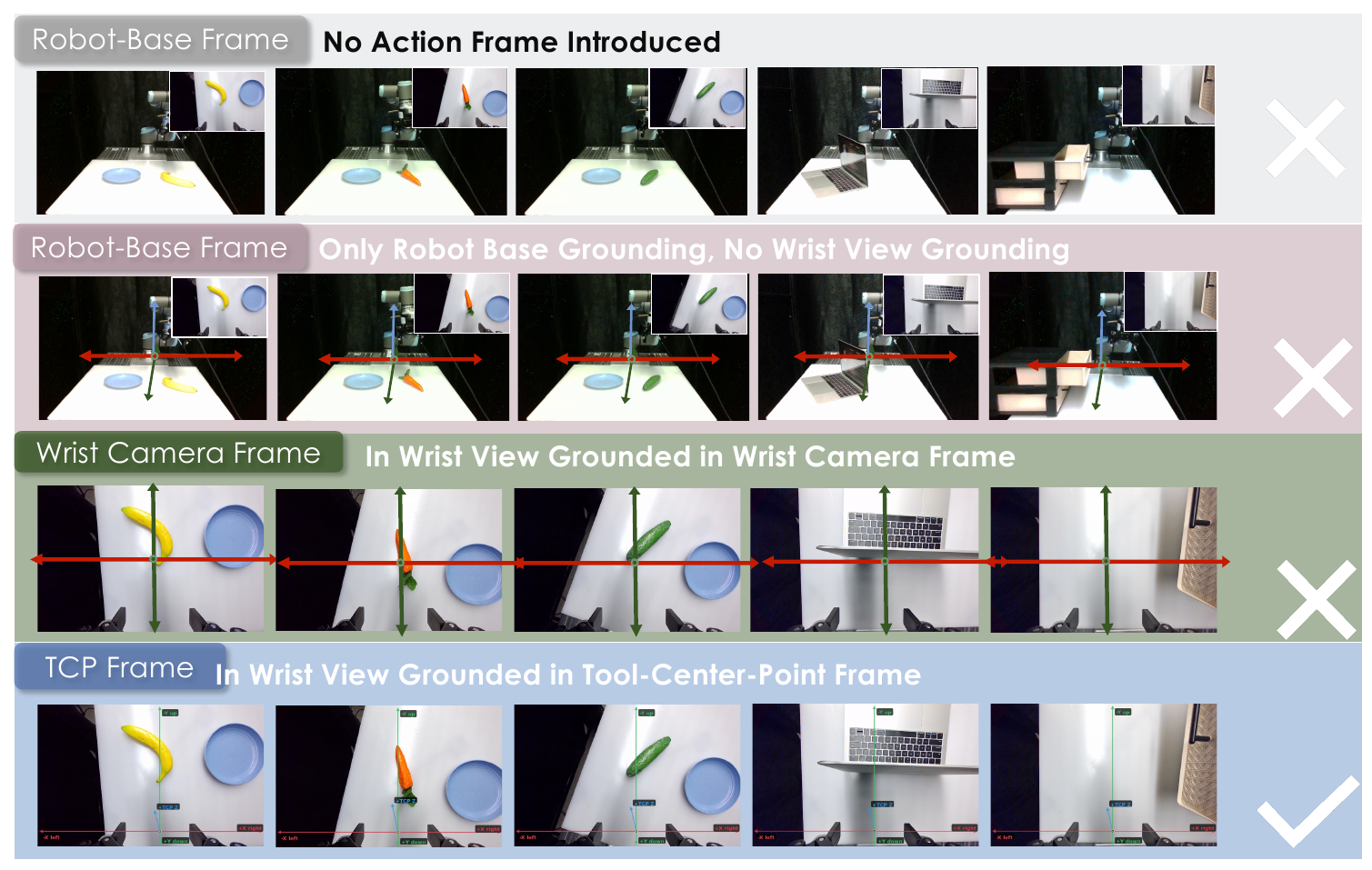}

    \caption{
    \textbf{Qualitative comparison of action-frame grounding.}
    We visualize representative rollouts of the full TCP-centered grounding and
    two alternative coordinate-frame choices on the same zero-shot manipulation
    task.
    }
    \label{fig:abl-action-frame}
\end{figure}

\end{document}